%% file: main.tex
\documentclass{article}
\PassOptionsToPackage{numbers,compress,sort}{natbib}
    \usepackage[final]{neurips_2022}

\usepackage[utf8]{inputenc} % allow utf-8 input
\usepackage[T1]{fontenc}    % use 8-bit T1 fonts
\usepackage{hyperref}       % hyperlinks
\usepackage{url}            % simple URL typesetting
\usepackage{booktabs}       % professional-quality tables
\usepackage{amsfonts}       % blackboard math symbols
\usepackage{nicefrac}       % compact symbols for 1/2, etc.
\usepackage{microtype}      % microtypography
\usepackage{xcolor}         % colors
\usepackage{bm}
\usepackage{pifont}% http://ctan.org/pkg/pifont
\usepackage{placeins}
\usepackage{enumitem} % added <<<<<<<<<<<<<

\usepackage[algo2e]{algorithm2e}

\usepackage{amsmath}
\usepackage{float}
\usepackage{graphicx}
\usepackage{caption}
\usepackage{subcaption}
\usepackage{multirow}
\usepackage{algorithm}
\usepackage{algcompatible}
\DeclareMathOperator*{\softmax}{softmax}

\newcommand{\cmark}{\ding{51}}%
\newcommand{\xmark}{\ding{55}}%

\newenvironment{talign*}
 {\csname align*\endcsname}
 {\endalign}

\input{std-macros}

\title{Test-Time Adaptation via Conjugate Pseudo-labels}

\author{%
  Sachin Goyal$^{\star 1}$ \quad Mingjie Sun$^{\star 1}$ \quad Aditi Raghunathan$^1$ \quad Zico Kolter$^{1,2}$\\
  $^1$Carnegie Mellon University, $^2$Bosch Center for AI \\ 
  \texttt{\{sachingo, mingjies, raditi, zkolter\}@cs.cmu.edu}
}
\begin{document}

\maketitle

\begin{abstract}
Test-time adaptation (TTA) refers to adapting neural networks to distribution shifts, with access to only the unlabeled test samples from the new domain at test-time. Prior TTA methods optimize over unsupervised objectives such as the entropy of model predictions in TENT~\citep{tent}, but it is unclear what exactly makes a good TTA loss. In this paper, we start by presenting a surprising phenomenon: if we attempt to \emph{meta-learn} the ``best'' possible TTA loss over a wide class of functions, then we recover a function that is \emph{remarkably} similar to (a temperature-scaled version of) the softmax-entropy employed by TENT. This only holds, however, if the classifier we are adapting is trained via cross-entropy loss; if the classifier is trained via squared loss, a different ``best'' TTA loss emerges.
To explain this phenomenon, we analyze test-time adaptation through the lens of the training losses's \emph{convex conjugate}.  We show that under natural conditions, this (unsupervised) conjugate function can be viewed as a good local approximation to the original supervised loss and indeed, it recovers the ``best'' losses found by meta-learning. This leads to a generic recipe that can be used to find  a good TTA loss for \emph{any} given supervised training loss function of a general class. Empirically, our approach consistently dominates other TTA alternatives over a wide range of domain adaptation benchmarks. Our approach is particularly of interest when applied to classifiers trained with \emph{novel} loss functions, e.g., the recently-proposed PolyLoss \citep{polyloss} function, where it differs substantially from (and outperforms) an entropy-based loss. Further, we show that our conjugate based approach can also be interpreted as a kind of self-training using a very specific soft label, which we refer to as the \emph{conjugate pseudo-label}. Overall, our method provides a broad framework for better understanding and improving test-time adaptation. Code is available at
\url{https://github.com/locuslab/tta_conjugate}.

\end{abstract}

\input{macros}
\input{introduction}

\input{Preliminary}
\input{metaloss}

\input{method_new}
% \input{method}
\input{experiment}
\input{related_works}
\input{conclusion}
\clearpage
\newpage 
\bibliography{ref_neurips22}
\bibliographystyle{plainnat}

\appendix
\include{appendix}

% \bibliography{ref_neurips22.bib} 
\end{document}

%% file: std-macros.tex
\newcommand\sL{\ensuremath{\mathcal{L}}}

\newcommand\sY{\ensuremath{\mathcal{Y}}}

\newcommand\R{\ensuremath{\mathbb{R}}} % Real numbers
\newcommand\eqdef{\ensuremath{\stackrel{\rm def}{=}}} % Equal by definition
\ifthenelse{\isundefined{\definition}}{\newtheorem{definition}{Definition}}{}
\ifthenelse{\isundefined{\assumption}}{\newtheorem{assumption}{Assumption}}{}
\ifthenelse{\isundefined{\hypothesis}}{\newtheorem{hypothesis}{Hypothesis}}{}
\ifthenelse{\isundefined{\proposition}}{\newtheorem{proposition}{Proposition}}{}
\ifthenelse{\isundefined{\theorem}}{\newtheorem{theorem}{Theorem}}{}
\ifthenelse{\isundefined{\lemma}}{\newtheorem{lemma}{Lemma}}{}
\ifthenelse{\isundefined{\corollary}}{\newtheorem{corollary}{Corollary}}{}
\ifthenelse{\isundefined{\alg}}{\newtheorem{alg}{Algorithm}}{}
\ifthenelse{\isundefined{\example}}{\newtheorem{example}{Example}}{}

%% file: macros.tex
\newtheorem{remark}{Remark}
\newcommand{\conjf}[1]{\ensuremath{{#1}^\star}}
\newcommand{\conjloss}[1]{\ensuremath{{#1}^\text{conj}}}
\newcommand{\ellxent}{\ell_\text{xent}}
\newcommand{\logsumexp}{\text{LogSumExp}}
\newcommand{\sent}{\text{softmax-entropy }}
\newcommand{\pseudolabel}{\text{pseudo-label }}
\newcommand{\pseudolabels}{\text{pseudo-labels }}
\newcommand{\ellsq}{\ell_\text{sq}}
\newcommand{\sourcetheta}{\theta_0}
\newcommand{\lsource}{\hat{L}_\text{source}}
\newcommand{\dsource}{D_\text{source}}
\newcommand{\dtest}{D_\text{test}}
\newcommand{\scaledpred}{\bar{h}}
\newcommand{\cpl}{\tilde{y}^\text{CPL}}

%sachin
\newcommand{\tta}{\text{TTA }}

%% file: introduction.tex
\section{Introduction}\label{Introduction}
Modern deep networks perform exceeding well on new test inputs that are close to the training distribution. However, this performance dramatically decreases on test inputs drawn from a different distribution. While there is a large body of work on improving the robustness of models, most robust training methods are highly specialized to the setting they cater to. For e.g., they assume pre-specified perturbations, subpopulations, and spurious correlations, or access to unlabeled data from the target distribution, and most methods offer close to no improvement on general distribution shifts beyond what they were trained for \citep{lost_dg, wilds}.
% tackling various aspects of 

% For instance, \citep{augmix} showed that even adding small amounts of ``natural'' noise like gaussian noise or snow and frost corruptions to the test data causes a steep decline in the test accuracy.  Training models robust to such distribution shifts \citep[e.g.]{groupDRO, coral, dann_adv} has traditionally required retraining models using unlabeled data from the new target domains.

In practice, it is often cumbersome (or even impossible) to precisely characterize all possible distribution shifts a model could encounter and then train accordingly. Instead, a model already trained on some source data must be able to \emph{adapt} at test-time to new inputs from a different domain. This setting of \emph{test-time adaptation} (TTA) has gained interest in recent years~\citep{tent,contrastive_tta,memo,ttt}. 
% However, owing to the impracticality of training new models each time distributions shift, there has been an increasing interest in \emph{test-time adaptation} \citep{tent,contrastive_tta,memo,ttt,bn_adapt}, which seeks to directly and quickly adapt pretrained models to distribution shift using unlabelled test data from the target domain.  
TTA is typically accomplished by updating the source model parameters via a few steps of optimization on an \emph{unsupervised objective} involving the new test sample from the target distribution. The choice of this unsupervised objective, which we call the TTA loss, dictates the success of the adaptation procedure.~\citep{ttt} uses a self-supervised objective on the test sample, \citep{tent} uses the entropy of model predictions, and several follow-ups have proposed variants or alternatives \citep{memo, rpl}. However, it remains unclear as to how to choose or guide the selection of this TTA loss, and thus far the choice of these losses has remained largely heuristic in nature.

In this work, we begin by presenting a set of intriguing experiments where we attempt to \emph{learn} the ``best'' TTA loss for a given source classifier and distribution shift. We parameterize the TTA loss by another neural network whose parameters are learnt via meta-learning~\citep{MAML,meta-loss} where we differentiate through the adaptation process to find the TTA loss that achieves the best adaptation on distribution shifts. Surprisingly, we ultimately learn a TTA loss that looks \emph{remarkably} similar to (a temperature-scaled version of) the softmax-entropy loss, which was already proposed by \citep{tent}. Why did we recover the commonly used softmax-entropy loss despite the fact that the procedure is capable of learning a very general class of losses and the meta-learning process could potentially \emph{specialize} to both the source classifier and the distribution shift of interest? Furthermore, we find that this pattern only holds when the loss used to train the source classifier is cross-entropy loss; when a different loss such as squared loss is used instead, the meta-learning procedure recovers a TTA loss that itself looks more like a negative squared error, and is very different from the softmax-entropy loss (Section ~\ref{sec:meta-loss}). 

In order to explain this phenomenon, we propose to consider \tta through the lens of the \emph{convex conjugate} function.  Specifically, given a hypothesis function $h(x)$ and label $y$, several common losses (cross-entropy and the squared loss amongst them, but not limited to these) can be written in the form $\mathcal{L}(h(x),y) = f(h(x)) - y^T h(x)$ for some function $f$.  In these cases, we show that ``natural'' TTA loss for such classifiers is precisely the (negation of) the convex conjugate evaluated at the gradient of $h$, $\mathcal{L}_{\mathrm{TTA}}(x) = -f^*(\nabla f(h(x))$, where $f^*$ is the convex conjugate of $f$. This framework not only recovers the results of our meta-learning experiments, but also justifies why some specific choices of TTA loss in the previous literature work well (e.g., this framework recovers TENT's choice of softmax-entropy for cross-entropy-trained classifier). Moreover, it also provides a broad framework for what the TTA loss should be when the source model is trained using various different loss functions (for example the recently-proposed PolyLoss \citep{polyloss, focalloss}) as is becoming increasingly common in machine learning.  Further, we show that our proposed conjugate adaptation loss is in fact a kind of self-training with \pseudolabels \citep{PL_paper}, a classic approach in machine learning. Various formulations of the \pseudolabel have been proposed in the literature, and our conjugate analysis provides a general recipe for the ``correct'' choice of soft pseudo-labels given by $\hat{y}(x) = \nabla f(h(x))$. We thus refer to these as \textit{conjugate pseudo-labels} (Conjugate PL's), and believe our work provides a broad framework for understanding adaptation with unlabeled data in general. 
%Interestingly, our experiments reveal that the learnt \tta loss for a source classifier trained using cross entropy loss indeed mimics something which is essentially a softmax-entropy (which \citep{tent, memo} heuristically proposed as a \tta loss). On the other hand, we observe that for a source classifier trained using squared loss, the learnt \tta loss mimics a quadratic function. These observations point towards a fundamental link between the loss used for training the source classifier and a possible best choice of \tta loss. We formalize this relation between the training and the \tta loss using our proposed framework based on convex conjugate analysis of the source classifier training.

%Our conjugate framework not only justifies why some specific choices of \tta loss in the previous literature work well (for example entropy in TENT), 
Finally, we empirically verify the effectiveness of our proposed conjugate adaptation loss across several datasets and training losses, such as cross-entropy and squared loss, along with the recently-proposed PolyLoss \citep{polyloss} (which itself has shown higher standard test accuracy on a wide range of vision tasks). Over \emph{all} models, datasets and training losses, we find our proposed conjugate pseudo-labeling consistently outperforms prior TTA losses and improves TTA performance over the current state of the art.
%Using our proposed test-time adaptation loss, which takes into account the source training loss function, the gains on standard accuracies using recently proposed loss functions can be further amplified on domain adaptation benchmarks, as we show in \autoref{sec:experiments}. Further, we find that for source classsifiers trained with cross-entropy loss, a simple temperature scaling of logits along with our proposed conjugate, which in this case would be entropy itself, outperforms or matches several heuristic follow ups to domain adaptation using entropy minimization (TENT, \citep{tent}).
%In summary, our paper reveals a fundamental link between optimal test-time adaptation loss and the source model training. We formalize this relation using our conjugate analysis framework and verify it empirically over multiple choices of training loss functions and datasets.

%% file: preliminary.tex
\section{Background and preliminaries.}\label{sec:preliminary}
% \subsection{Preliminary and setup}
\textbf{Test-time adaptation.} We are interested in mapping an input $x \in \R^d$ to a label $y \in \sY$. We learn a model $h_\theta:\R^d \mapsto \R^{| \sY |}$ parameterized by $\theta$ that maps an input $x$ to predictions $h_\theta(x)$. 
We assume access to a trained source model and adapt at test-time over the test input, before making the final prediction. This is the standard test-time adaptation (TTA) setting~\citep{ttt,tent}. During TTA, we update the model parameters on an unsupervised objective $\mathcal{L}(x, h_{\theta})$. For example, in TENT~\citep{tent}, this loss is the entropy of the softmax-normalized predictions of the model. At each time step of adaptation, we observe a batch of test inputs and we take a gradient step towards optimizing the TTA loss on this test batch. As is standard, we measure the average online performance of models across all steps (number of test batch inputs seen) in the adaptation process. 
% For example, the objective in TENT~\citep{tent} is $\mathcal{L}_\text{tent}(h_\theta(x))= \mathcal{H}(\text{softmax}(z))$, where $z=h_{\theta}(x_t)$ is the model prediction and $\mathcal{H}$ is the entropy function. With online adaptation, model is updated while predicting test data at the same time and average performance throughout testing is measured.

\textbf{Meta learning the loss function.} In order to explore the existence of different TTA losses, we employ the meta-learning procedure where we attempt to \emph{learn} the TTA loss. We use a similar procedure as prior work on meta-learning loss functions~\citep{meta-loss,discover-rl} and parameterize the loss function via a neural network $m_{\phi}:\R^{|\sY|} \mapsto \R$ that takes in the model predictions/logits and outputs a loss value. We want to learn parameter $\phi$ such that when we update $\theta$ via the loss function $m_{\phi}$, our final performance is optimal. In order to do so, let $x$ be the unlabeled test samples to adapt to, and $y$ be the corresponding labels. We update $\theta$ and $\phi$ alternatively as follows. 
\begin{align}
\label{eq:metaupdate}
\theta^{t+1} \leftarrow\theta^t-\alpha\frac{\partial m_{\phi^t}(h_{\theta^{t}}(x))}{\partial \theta^t},~~ 
\phi^{t+1} \leftarrow \phi^t - \beta\frac{\partial \mathcal{L}(h_{\theta^{t+1}}(x'), y')}{\partial \phi^t},
\end{align}
where $\mathcal{L}$ is some supervised surrogate loss function such as cross-entropy. Please refer to Appendix A3 for further details regarding meta-learning setup. Note that the meta-learning process above assumes access to labels $y$ of test inputs. In this paper, we do \emph{not} propose meta-learning the TTA loss as an approach. Rather, we use meta-learning to explore what the ``best'' TTA losses look like. We discuss our findings from this exploration in the next section. 

%% file: metaloss.tex
\section{Test-time Adaptation via Meta-Learnt Losses}\label{sec:meta-loss}
The objective used in TENT is the \sent of the model predictions which essentially makes the classifier more confident in its current predictions.
% However, there is no fundamental explanation as for why it works well in practice. 
% In principle, entropy minimization tries to assign a higher probability to the predicted class, 
The same can be achieved by various other loss formulations such as those mentioned in \citep{rpl}. With so many possible choices for the loss function, what should we use for TTA? In this section, we attempt to answer this empirically and present some intriguing observations. 
% However, there is a lack of a principled approach to guide or choose the selection of this \tta loss.% Similar arguments can be made for other self-learning based losses, e.g. pseudo labelling~\citep{rusak2022if}. 
\begin{figure}[!htbp]
	\centering
	\begin{subfigure}{0.40\textwidth}
		\includegraphics[width=\textwidth]{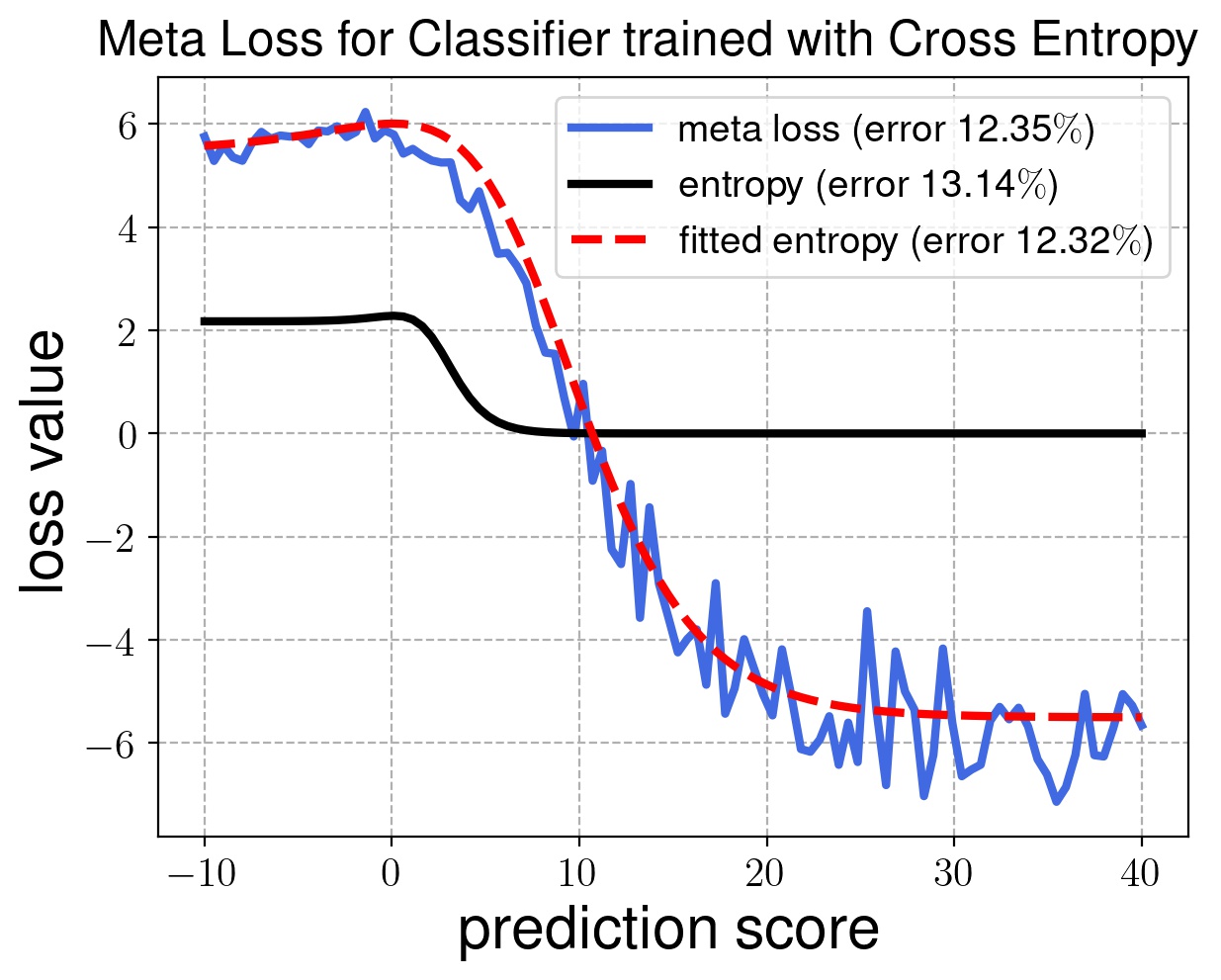}
% 		\vspace{-1ex}
		\caption{}
		\label{fig:meta-loss-ce-trained}
	\end{subfigure}
%%%%%%%%%%%%%%
	\begin{subfigure}{0.40\textwidth}
		\includegraphics[width=\textwidth]{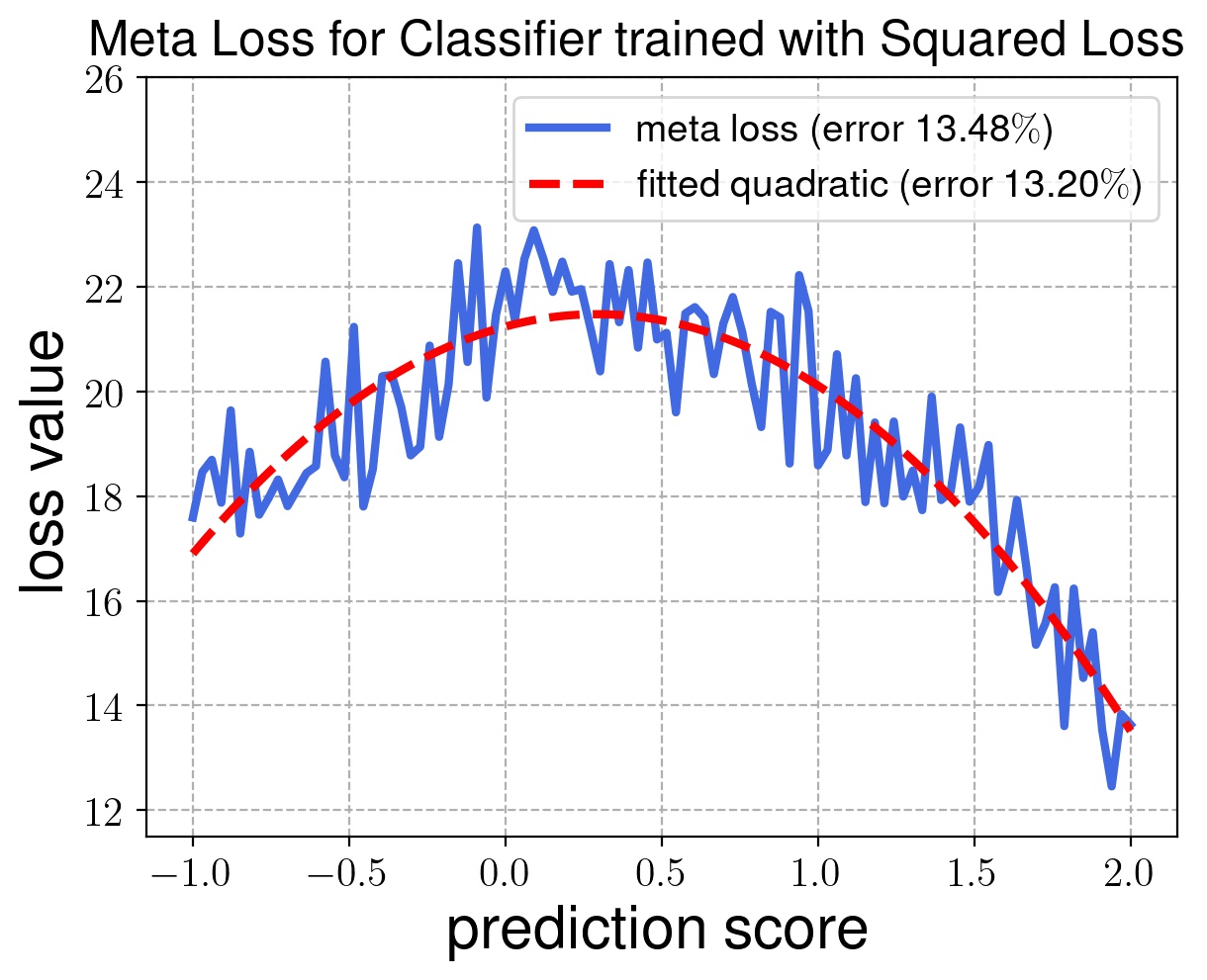}
% 		\vspace{-1ex}
		\caption{}
		\label{fig:meta-loss-quadratic}
	\end{subfigure}
	\vspace{-1ex}
  \caption{Visualization of meta loss (blue) by varying one input prediction score. (a) For cross-entropy loss trained model, the learnt meta loss can be approximated with a scaled softmax-entropy function (dashed red). (b) When the source model is trained with a squared loss for classification, the learnt meta loss (blue) can be fitted closely with a quadratic function (dashed red), shown in~\autoref{fig:meta-loss-quadratic}. The range (max/min) of the prediction score (logit) in x-axis is chosen to cover the empirical range of the predicted logits.}
\label{fig:fitted_ce}
\end{figure}
% \vspace{-1.5em}
\paragraph{Experiment 1.} We learn the TTA loss parameterized by a neural network via meta-learning as described in Section~\ref{sec:preliminary}. Our source classifier is a ResNet-26 trained on CIFAR-10 and we adapt to distribution shifts in CIFAR-10-C. 
% We focus our attention on learning a meta-loss from the family of functions over the model predictions $z=h_{\theta}(x)$ i.e. $\{l(z) : \mathbb{R}^k \rightarrow \mathbb{R}\}$, where $k$ is the number of categories in the classification task.
% Consider the task of test time adaptation (TTA) from CIFAR-10 to CIFAR-10-C corruptions dataset~\citep{corruptions_data}. We take a ResNet-26 trained on the source data i.e. CIFAR-10 using cross entropy loss as our source classifier $h_{\theta}$. As described in \autoref{sec:preliminary}, we try to learn a meta-loss parameterized by a neural network $m_{\phi}$ over the model predictions $h_{\theta}(x)$, such that minimizing the loss $m_{\phi}(h_{\theta}(x_t))$ at test time improves generalization on the target domain. 
We use the 4 labeled validation noises in CIFAR-10-C to learn the meta-loss network parameters and we denote the resulting learnt loss function by meta-TTA loss. We then adapt the source classifier to the test set of 15 corruptions by optimizing the meta-TTA loss. 
% The meta-loss $m{\phi}$ is learnt using the 4 validation noises in the CIFAR-10-C dataset. The learnt meta-loss is then used as a \tta loss while making predictions over the test set of 15 noises with highest level of severity. The source model parameters $\theta$ are updated at test time as specified in \autoref{sec:preliminary}.
\paragraph{Observations.}
First, we find that TTA using meta-TTA loss performs better than TENT ($12.35\%$ vs $13.14\%$), suggesting that there are better TTA losses than previous losses based on softmax-entropy. 

However, on examining this meta-TTA loss, we find a surprising observation. \autoref{fig:meta-loss-ce-trained} (blue curve) visualizes the learnt meta-loss over model predictions as we vary a single class prediction with the rest fixed. Qualitatively, the learnt meta-loss looks very similar to softmax-entropy in one dimension. In fact, we can fit it closely with a scaled softmax-entropy function (dashed red curve): $\alpha\cdot \mathcal{H}(\text{softmax}(h_{\theta}(x)/T))$, where $\alpha$ is a magnitude parameter and $T$ is a temperature scaler. We want to test if the meta-loss is basically learning the softmax-entropy function. Hence, we perform test-time adaptation with the fitted softmax-entropy function instead (dashed red curve) and achieve an error of $12.32\%$, essentially recovering the performance of meta-TTA.
% an extra temperature scaling before softmax normalization (red curve). 
% In fact, if we just fit a scaled entropy function  directly to the meta loss in~\autoref{fig:meta-loss-ce-trained} and perform TENT with the fitted entropy function, we achieve an error of $12.26\%$ and hence essentially recover the performance of meta-TTA.

Despite the ability to represent many different loss functions and potentially specialize to the CIFAR-10-C setting, the meta-loss procedure gave back the standard entropy objective. \emph{Do we always recover a loss that looks like softmax-entropy?}
% can be closely approximated by a scaled version of softmax entropy i.e. $\alpha\cdot \mathcal{H}(\text{softmax}(x/T))$ (\autoref{fig:meta-loss-ce-trained}, dashed red curve). 
% Further, this scaled entropy function achieves a mean error of 12.26$\%$ which matches the performance of meta loss for \tta on CIFAR-10-C. We find that even varying the $\mathcal{L}_\text{task}$ in \autoref{sec:preliminary} to other loss functions gives similar meta-loss. In summary, it turns out that we started off trying to find a better function than entropy via learning a function over a much more powerful class of $M_{\phi}$ but still ended up re-discovering the entropy function yet with a temperature scaling, even though there was no explicit supervision of entropy signal during meta-loss learning.
\paragraph{Experiment 2.} In an attempt to isolate when we get back the entropy objective, we vary several things. We tried different architectures for the source classifier, different losses $\sL$ during the meta-learning process~\eqref{eq:metaupdate} and different training losses for the source classifier.

\paragraph{Results.}
We observed that we consistently recovered the temperature scaled softmax-entropy function in all cases \emph{except} when we varied the training loss for the source classifier (Appendix ~\ref{app:additional_meta_learn}). On using the squared loss function~\citep{sq_loss_paper}, a strikingly different meta-TTA loss emerges. ~\autoref{fig:meta-loss-quadratic} (blue curve) shows the learnt meta-loss (13.48$\%$ error) for this network. Here again, the meta-TTA loss outperforms entropy ($14.57\%$) but it is not simply due to a scaling factor. The loss now looks like the negative squared error (red curve). Like previously, we tried fitting a quadratic loss directly to the meta loss in~\autoref{fig:meta-loss-quadratic}, and this time we even slightly outperformed the meta-TTA loss. 

% that outperforms the error when softmax-entropy (14.57$\%$) is used as \tta loss over the squared loss trained classifier . Interestingly, here the learnt meta-loss can be approximated by a quadratic function (~\autoref{fig:meta-loss-quadratic}, dashed red curve) which performs even better (13.20$\%$ mean error) as compared to the meta loss and entropy .

To summarize, we used a meta-learning procedure to search for the ``best'' TTA loss, where the loss itself was parameterized by a neural network that could potentially represent arbitrarily complex loss functions. However, we ended up with loss functions displaying remarkable structure: across different architectures and different variants of meta-learning, for a classifier trained with cross-entropy, the meta-TTA loss was temperature scaled softmax-entropy and for a classifier trained with squared loss, the meta-TTA loss was a negative squared loss.   This is interesting from both a practical and conceptual standpoint where the ``best'' TTA loss depends on the loss used to train the source classifier in a clean fashion. We attempt to understand and explain this phenomenon in the next section. 
% The above experiments suggest that the \textit{choice of test time adaptation objective should take into consideration the supervised learning objective used for training the source model}. In the next section, we formalize this by explicitly modelling the relation between the source training loss function and what should be the choice of test time adaptation objective. 

%% file: method_new.tex
% \vspace{-1em}
\section{Conjugate Pseudo Labels}\label{sec-conjugate}
% \vspace{-0.5em}
Results in the previous section raise an obvious question: \emph{why} does softmax-entropy as used in TENT seem to be the ``best'' possible test time adaptation loss for classifiers trained via cross-entropy (at least, best in the sense that meta-learning consistently recovers something which essentially mimics softmax-entropy, even though meta-loss is parameterized by a neural network and hence could learn much more complex functions specific to the model and the particular shift)?  And why, alternatively, does a quadratic TTA loss seem to perform best when the classifier is trained via squared loss?

In this section, we offer an explanation of this phenomenon via the construct of the convex conjugate function \citep{conj_wiki}.  As we will see, our method recovers softmax-entropy \emph{and} quadratic loss as the ``natural'' objectives for classifiers trained via cross-entropy and squared loss respectively.  Furthermore, for classifiers trained via \emph{other} loss functions, as is becoming increasingly common in deep learning, our approach naturally suggests corresponding test-time adaptation losses, which we show in the next section to comparatively outperform alternatives.  Thus, we argue that our framework overall provides a compelling recipe for specifying the ``correct'' method for \tta for a large class of possible losses.

% \vspace{-0.5em}
\subsection{Losses and the convex conjugate}

We begin by formally considering loss functions between a hypothesis output $h_\theta(x)$ (e.g., the logit outputs of a classifier, or the direct prediction of a regressor) and target $y$ that take the following form
\begin{equation}
\label{eq:loss_separable}
    \mathcal{L}(h_\theta(x),y) = f(h_\theta(x)) - y^T h_\theta(x)
\end{equation}
for some function $f$; when there is no risk of confusion, we will use $h$ in place of $h_\theta(x)$ for simplicity of notation. While not every loss can be expressed in such a form, this captures a wide variety of common losses (possibly scaled by a constant value).  For example, cross-entropy loss corresponds to the choice $f(h) = \log \sum_i \exp(h_i)$ and where $y$ denotes a one-hot encoding of the class label; similarly, squared loss corresponds to the choice $f(h) = \frac{1}{2} \|h\|_2^2$.

When training an over-parameterized classifier, we can roughly view the training process as (approximately) attaining the \emph{minimum} over hypotheses $h$ for each training example
\begin{equation}
    \min_\theta \frac{1}{t} \sum_{i=1}^t \mathcal{L}(h_\theta(x_i), y_i) \approx \frac{1}{t} \sum_{i=1}^t \min_h \mathcal{L}(h,y_i)
\end{equation}
where $t$ is the number of training samples. However, in the case of losses in the form \eqref{eq:loss_separable}, the minimization over $h$ in this form represents a very specific and well-known optimization problem: it is known as the \emph{convex conjugate} \citep{conj_wiki} of the function  $f$
\begin{equation}
    \min_h \mathcal{L}(h,y) =  \min_h \{f(h) - y^T h \} = -f^\star(y)
\end{equation}
where $f^\star$ denotes the convex conjugate of $f$. $f^\star$ is a convex function in $y$ (and indeed, is convex regardless of whether or not $f$ is convex).  Furthermore, for the case that $f$ is convex differentiable, the optimality condition of this minimization problem is given by $\nabla f(h^\text{opt}) = y$, so we also have that
\begin{equation}
    f^\star(y) = f^\star(\nabla f(h^\text{opt}))
\end{equation}
where $h^\text{opt}$ refers to the optimal classifier (used interchangeably with $h_{\theta^\text{opt}}$). Putting this all together, we can state (admittedly, in a rather informal manner) that under the assumption that $\theta^\text{opt}$ is chosen so as to approximately minimize the empirical loss on the source data in the over-parameterized setting, we have that for $t$ inputs
\begin{equation}
    \frac{1}{t} \sum_{i=1}^t \mathcal{L}(h_{\theta^\text{opt}}(x_i), y_i) \approx \frac{1}{t} \sum_{i=1}^t -f^\star(\nabla f(h_{\theta^\text{opt}}(x_i)))
\end{equation}
i.e., the empirical loss can be approximated by the (negative) conjugate applied to the gradient of the $f$, at least in a region close to the optimal $\theta^\text{opt}$ that minimizes the empirical loss.  But the later expression has the notable benefit that it does not require any label $y_i$ in order to compute the loss, and thus can be used as a basis for \tta on target domain of the hypothesis function $h_{\theta^\text{opt}}$. 

\begin{definition}[conjugate adaptation loss]
\label{def:conjugateadapt}Consider a loss function that takes the form given in \ref{eq:loss_separable}, used for training a hypothesis $h_{\theta}$ in the over-parameterized regime. We define the conjugate adaptation loss $\mathcal{L}^\text{conj}(h_{\theta}(x)): \R^{|\sY|} \mapsto \R$ as follows. 
\begin{align}
\label{eqn:conjugateadapt}
    \mathcal{L}^\text{conj}(h_{\theta}(x)) = -\conjf{f}(\nabla f(h_{\theta}(x))) = f(h_{\theta}(x)) -  \nabla f(h_{\theta}(x))^\top h_{\theta}(x).
\end{align}
\end{definition}

\subsection{Recovery of existing test-time adaptation strategies} 
\label{sec:recovery}
\paragraph{Cross-entropy}The interesting aspect to this formalism is that when applied to classifiers trained with cross-entropy, it recovers exactly the TENT approach to \tta: minimizing the softmax-entropy of $h_\theta(x)$.  And indeed, this loss was \emph{also} recovered when using meta-learning to learn the ``optimal'' test-time adaptation loss. To see this, note that for cross-entropy, we have that $f(h) = \log\sum_i \exp(h_i)$, giving the optimality condition $y = \nabla f(h^\text{opt}) = \frac{\exp(h^\text{opt})}{\sum_i \exp(h_i^\text{opt})}$ and the conjugate function
\begin{equation}
    f^\star(y) = \left \{ \begin{array}{ll} \sum_i y_i \log y_i & \mbox{ if } \sum_i y_i = 1 \\
    \infty & \mbox{ otherwise} \end{array} \right . .
\end{equation}
In other words, 
\begin{equation}
\mathcal{L}^\text{conj}(h_{\theta}(x)) = -f^\star(\nabla f(h_\theta(x))) = - \sum_i \frac{\exp(h_i)}{\sum_j \exp(h_j)} \log \frac{\exp(h_i)}{\sum_j \exp(h_j)}
\end{equation}
i.e. softmax-entropy of the model prediction, which is exactly the \tta loss that TENT uses.

\paragraph{Squared loss} \label{sec:cpl_sqloss}
For the squared loss, we have that $f(h) = \frac{1}{2} \|h\|_2^2$, leading to the optimality condition $y = h$ and conjugate function $f^\star(y) =\frac{1}{2} \|y\|_2^2$. Hence, the adaptation loss in this case would be simply given by $\mathcal{L}^\text{conj}(h_{\theta}(x)) = -f^\star(\nabla f(h_\theta(x))) = -\frac{1}{2}\|h\|_2^2$ which is also what we observed in the meta-learning experiments discussed in Section ~\ref{sec:meta-loss}.

%The use of a negative quadratic adaptation loss when training under squared error (which we also observed in the meta-learning setting) is also well-motivated by this conjugate formulation.  In this case we have that $f(h) = \frac{1}{2} \|h\|_2^2$, leading to the optimality condition $y = h$ and conjugate function $f^\star(y) =\frac{1}{2} \|y\|_2^2$.  Thus, the adaptation loss in this case would simply by given by
%\begin{equation}
    % -f^\star(\nabla f(h_\theta(x))) = -\frac{1}{2}\|h\|_2^2.
%\end{equation}

\subsection{Conjugate pseudo-labels}
\label{sec:cpl}
We now emphasize that by the nature of our approximations, there is an additional simple interpretation of the conjugate loss: it is also equal to the original loss (\ref{eq:loss_separable}) applied to the ``psuedo-labels'' $\cpl_\theta(x) = \nabla f(h_\theta(x))$, where CPL refers to conjugate pseudo-labels, i.e.,
\begin{equation}
    \mathcal{L}^\text{conj}(h_{\theta}(x)) = -f^\star(\nabla f(h_\theta(x))) = f(h_\theta(x)) - \nabla f(h_\theta(x))^T h_\theta(x) = \mathcal{L}(h_\theta(x), \nabla f(h_\theta(x))).
\end{equation}
This property is known as the Fenchel-Young inequality, that is $f(x) + f^\star(u) \geq x^T u$ holding with equality when $u = \nabla f(x)$.  In other words, our conjugate adaptation loss is precisely equivalent to self-training under the specific soft pseudo-labels given by $\cpl = \nabla f(h_\theta(x))$.  And indeed, for many cases, this may be a more convenient form to compute than explicitly computing the conjugate function at all. For this reason, we refer to our method as that of \emph{conjugate pseudo-labels}.

In the case of cross-entropy loss, this approach then corresponds exactly to self-training using labels given by the softmax applied to the current hypothesis.  We must emphasize, however, that while our conjugate formulation indeed has this ``simple'' form for the case of cross-entropy loss, the real advantage comes in that it provides the ``correct'' \pseudolabel for use with \emph{other} losses, which may result in pseudo-labels different from the ``common'' softmax operation.

\paragraph{Example: conjugate pseudo-labels for PolyLoss.}
\label{sec:cpl_poly}
PolyLoss \citep{polyloss} is a recently-proposed simple alternative to cross-entropy loss than has been shown to improve performance across a wide variety of compute tasks. This loss is given by the form
\begin{equation}
    \mathcal{L}_{\mathrm{poly}}(h_\theta(x),y) = \mathcal{L}_{\mathrm{ce}}(h_\theta(x),y) + \epsilon \cdot y^T(1 - \softmax(h_\theta(x)))
\end{equation}
We note that this can be put exactly into our conjugate form (equation \ref{eq:loss_separable}) by writing the loss in a slightly more involved fashion, which we refer to as the \emph{expanded conjugate form}
\begin{equation}
    \mathcal{L}_{\mathrm{poly}}(h_\theta(x),y) = f(h_\theta(x)) - y^T g(h_\theta(x)).
\end{equation}
where $f$ is the log-sum-exp function as before, and $g(h)=h - \epsilon(1-\softmax(h))$.  In order to formally put this into the form of the previous loss function (equation \ref{eq:loss_separable}), we can simply define an alternative hypothesis as the function $h_\theta'(x) = g(h_\theta(x))$, and then define PolyLoss in the conjugate form as
\begin{equation}
    \mathcal{L}_{\mathrm{poly}}(h'_\theta(x),y) = f(g^{-1}(h'_\theta(x))) - y^T h'_\theta(x).
\end{equation}
Typically, however, it is easier to simply operate on the expanded conjugate form, which yields the optimality condition for the  pseudo-label $\nabla f(h^\text{opt}) = \mathsf{D} g(h^\text{opt}) \cpl_\theta(x)$, where $\mathsf{D}$ is the Jacobian operator. For the case of PolyLoss, this leads to the conjugate pseudo-label of the following form: $\cpl_\theta(x)  = (I + \epsilon \mathrm{diag}(z) - \epsilon z z^T)^{-1}z, \;\; z \equiv \softmax(h_\theta(x))$.
% \vspace{-2ex}
% \begin{equation}
%     \label{eq:conjugate_poly}
%     \cpl_\theta(x)  = (I + \epsilon \mathrm{diag}(z) - \epsilon z z^T)^{-1}z, \;\; z \equiv \softmax(h_\theta(x)).
% \end{equation}
%Although this inverse may be impractical to compute directly when the number of classes is large, it can still easily be computed in these cases via the low-rank matrix inversion lemma.  More generally, for any function where $D(g(h_\theta(x)))$ is non-singular, the conjugate psuedolabel based upon an extended conjugate loss allows for a simple computation of the corresponding conjugate \pseudolabel.

%\subsection{Conjugate \pseudolabels for test-time adaptation}

\paragraph{Test-time adaptation.} Finally, we note that the above discussion doesn't actually address any topics related to test-time adaptation to OOD data, but merely provides a generic characterization of a self-training procedure for generic loss functions of the form \eqref{eq:loss_separable}.  However, the application to  \tta on OOD data is fairly straightforward: as long as the learnt source parameters $\theta$ is a reasonable approximation to the true optimal $\theta^\text{opt}$ on the shifted domain, self-training with the conjugate \pseudolabels provides a reasonable proxy for fine-tuning the network on the true OOD loss. We emphasize that, common to most approaches for \tta, there are still some amount of design decisions that must be put in place; these are detailed in Section \ref{sec:imp_details}. In practice, we observe OOD generalization typically benefits (across all baselines) from an additional “temperature” scaling, i.e., applying the TTA loss to $h_{\theta}(x)/T$ for some fixed temperature $T$, although it requires a held-out validation dataset for tuning $T$. However, we should emphasize that truly unsupervised \tta would require making an informed guess for the value of these hyper-parameters. The full procedure for test time adaptation via conjugate pseudo-labels is shown in Algorithm \ref{alg:conjPL}.

\begin{algorithm}
  \caption{Conjugate pseudo-labeling (Conjugate PL)}
  \label{alg:cpt}
  \label{alg:conjPL}
    \hspace*{\algorithmicindent}
    \textbf{Input:} Source classifier $\sourcetheta$ trained using loss $\mathcal{L}(h_\theta(x), y) = f(h_\theta(x)) - h_\theta(x)^\top y$.\\
    \hspace*{\algorithmicindent} \hspace*{\algorithmicindent} \hspace*{\algorithmicindent} $N$ batches of test data $\dtest = [x_1, x_2, \hdots, x_N]$
    %$\mathcal{L}_\text{source}(\theta) = \sum\limits_{(x, y) \in \dsource} \ell( h_\theta(x), y)$, where $\ell(z, y)=f(z) - y^\top z$. \textbf{hparams:}
    \\
    \hspace*{\algorithmicindent} \textbf{Hyperparams:}
    learning rate $\eta$ and temperature $T$.
  \begin{algorithmic}[1]
    \STATEx 
    Let $\scaledpred_\theta(x) \eqdef h_\theta(x)/T$ be the temperature scaled predictor. 
    \STATEx
    \vspace{4pt}
    Let $\cpl_\theta(x)$ denote the conjugate pseudo-label function $\cpl_\theta(x) =\nabla(f(\bar{h}_\theta(x)))$.
    \STATEx
    \vspace{4pt}
    \textbf{for} $n = 0, 1, \hdots N-1$ \textbf{do}
    \STATEx
    % \hspace{10pt} Sample $x_n \sim \dtest$.
    % \STATEx
    \hspace{10pt}
    \vspace{-1ex}
    \STATEx
    \hspace{10pt} $\theta_{n+1} = \theta_n - \eta \nabla \mathcal{L}\Big(\scaledpred_\theta(x_n), \cpl_\theta(x_n)\Big)$ ~~\text{[Self-training with conjugate pseudo-labels]}
  \end{algorithmic}
\end{algorithm}

%% file: experiment.tex
% \vspace{-0.5em}
\section{Experiments}\label{sec:experiments}
% \vspace{-0.5em}
In this section, we empirically evaluate the effectiveness and generality of the proposed conjugate pseudo-labeling procedure (Algorithm~\ref{alg:conjPL}) for test-time adaptation on a variety of datasets. 
% \vspace{-0.5em}
\subsection{Setup}
\paragraph{Datasets.} We evaluate on the three common corruption benchmarks: adapting a classifier trained on CIFAR-10 to CIFAR-10-C, CIFAR-100 to CIFAR-100-C and ImageNet to ImageNet-C~\citep{corruptions_data}. Following the previous works \citep{tent, ttt}, we report the error averaged across corruptions at the highest severity for CIFAR-10/100-C and averaged across corruptions and severity level for ImageNet-C.
We also evaluate on three domain adaptation datasets: adapting a classifier trained on SVHN to MNIST, an ImageNet classifier
to ImageNet-R~\citep{hendrycks2021many} and adapting from synthetic to real data in VISDA-C~\citep{visda2017}. 

\paragraph{Models and Training losses.} Following previous works on TTA\citep{tent,ttt}, we use ResNet-26 \citep{resnet} as the source classifier architecture for CIFAR-10/100 experiments, ResNet-18 for SVHN to MNIST and a ResNet-50 for ImageNet and source synthetic data on VisDA-C. We consider source classifiers trained via the following loss functions: the de-facto cross-entropy,  recently proposed polyloss \citep{polyloss} and squared loss \citep{sq_loss_paper}.

\paragraph{Baselines.} 
Our proposed conjugate pseudo-label is the classic approach of self-training with a specific form of pseudo-labels. In self-training, we replace the label $y$ with a pseudo-label $\tilde{y}(x)$ and adapt by optimizing the loss function $\mathcal{L}(h_\theta(x), \tilde{y}(x))$. Note that we could either instantaneously update the pseudo-labels using the current classifier, or generate pseudo-labels once with just the source classifier. Instantaneous updates have been shown to work better for domain adaptation~\citep{rpl, self_spurious}, and we perform instantaneous updates for all methods. 
While we propose using $\tilde{y}^\text{CPL}(x) = \nabla f(h_\theta(x))$ (See Section~\ref{sec:cpl}), we compare to the standard pseudo-labels used in the literature: 
\begin{itemize}[leftmargin=*]
\vspace{-5pt}
\itemsep0em 
\item (i) the ``hard'' pseudo-label (hard PL) where $\tilde{y}(x) = \arg\max_i \big(h_\theta(x)\big)_i$ is the most likely class as predicted by $h_\theta$. As is common in the self-training literature, we perform confidence thresholding.
\item (ii) The ``soft'' pseudo-label (soft PL) where $\tilde{y}(x)$ is obtained by applying a softmax function to the model predictions $h_\theta(x)$.
\end{itemize}
\vspace{-2.5pt}
We also compare with the following recently proposed test-time adaptation methods.
\begin{itemize}[leftmargin=*]
\vspace{-5pt}
\itemsep0em 
\item Entropy Minimization (ENT)~\citep{tent}  minimizes the entropy of model predictions.
\item Robust Pseudo-Label~\citep{rpl} where we minimize a robust classification loss,  $\mathcal{L}_\text{rpl}=q^{-1}(1-p(i|x)^q)$ where $i=\text{argmax}_jp(j|x)$ and $q\in[0,1]$.
\item MEMO~\citep{memo} minimizes entropy of a model's outputs across different augmentations of a test input. We implement a \textit{batch version}, where we see multiple test points at once, for fair comparisons.
\end{itemize}

\paragraph{TTA methodology.}\label{sec:imp_details} Following ~\citep{tent} and~\citep{rpl}, we fine-tune by updating the learnable scale and shift parameters of the batch normalization layers across all adaptation losses. For each batch, batch normalization statistics is also updated, as suggested in~\citep{bn_adapt}.
% It is also empirically shown~\citep{rusak2022if} to be better than updating all model parameters. Hence we update only the scale and shift parameters across all adaptation losses. 
We report performance at the end of one round of test-time adaptation over the entire test set. 

We tune the learning rate (LR) and temperature (T) on the validation noises in the corruption benchmark by grid-search. LR is selected from $\{1e^{-1}, 1e^{-2}, \hdots 1e^{-4}\}$ and T from $\{1,2 \hdots 5\}$. All the experiments have been performed on A6000 GPU's. On domain adaptation benchmarks, where there is no held-out target domain, we set T to be $1$ and use the LR suggested by \cite{tent,contrastive_tta}. We use the same hyperparameter tuning protocol across all methods. We single out temperature as a very important hyperparameter, as we discuss in the results below.  
\input{tables/table-ce}
\subsection{Results on classifiers trained with cross-entropy}
We study the effectiveness of our proposed conjugate pseudo-labels when the source classifier is trained via cross-entropy loss. In this case, baselines Softmax PL and ENT are the same as Conjugate PL. Thus we omit them in our results. Table~\ref{tab:ce_loss}, reports the performance of various TTA methods. 

When the source classifier is trained via cross-entropy, our conjugate pseudo-label algorithm exactly corresponds to entropy minimization with an additional temperature scaling. Entropy minimization as proposed in prior work~\citep{tent} does not tune the temperature parameter, and some newer objectives such as robust PL or MEMO outperform vanilla entropy minimization. For example, on CIFAR-100-C, vanilla ENT obtaines $41.15\%$ average error, while robust PL improves this to $39.80\%$ and MEMO to $38.52\%$. However, with the right temperature scaling, entropy minimization obtains $36.10\%$ error which outperforms the newer objectives (with and without temperature scaling). A similar observation holds for CIFAR-10-C and ImageNet-C as well. Essentially, the gains over vanilla entropy minimization vanish when we do temperature scaling, and entropy minimization (i.e. conjugate pseudo-labeling corresponding to cross-entropy) turns out to be the best objective after all. 

\subsection{Results on classifiers trained with polyloss and squared loss}
In the case of cross-entropy, conjugate pseudo-labeling reduces to the familiar notion of entropy minimization. We now explore the performance of our method on different loss functions where the conjugate pseudo-labels differ substantially from entropy minimization (\autoref{sec:cpl_poly}). Table~\ref{tab:polyloss1} presents the results on the corruption benchmarks and Table~\ref{tab:polyloss2} presents the results on the other domain adaptation datasets for source classifiers trained with PolyLoss.
\input{tables/table-polyloss1}

\input{tables/table-polyloss2}

% \vspace{-5mm}
 First, we note that, across all datasets in Table~\ref{tab:polyloss1} and Table~\ref{tab:polyloss2}, our conjugate PL approach outperforms all other \tta losses. With polyloss classifiers, entropy minimization is no longer the best method---on CIFAR-100-C, entropy minimization achieves $38.23\%$ error while our conjugate PL achieves $36.83\%$. We see similar \emph{consistent gains} on CIFAR-10-C, ImageNet-C, ImageNet-R and VisDA-C. On digit adaptation tasks from SVHN to MNIST/USPS/MNISTM, where there is a larger shift between source and target, the gains are especially pronounced. Figure~\ref{fig:task_loss_curve} compares how the task loss (polyloss $\epsilon=6$) on the test data decreases as we adapt the model through conjugate PL and other baselines.  We use CIFAR-10-C as an example. Observe that our proposed conjugate PL indeed reduces the task loss the most among other baselines.
 
 Furthermore, on CIFAR-10-C and ImageNet-C, we find that adapting polyloss classifiers via conjugate PL improves the performance over all methods applied to cross-entropy trained source classifiers. For e.g., on ImageNet-C, the performance improves from $45.34\%$ to $44.01\%$. However, this is only true when using the proposed conjugate PL. If we just did \sent minimization (even with temperature scaling), the final adapted performance of a polyloss classifier ($45.5\%$) is in fact worse than that of a cross-entropy classifier ($45.34\%)$. Our results suggest that as we develop new training losses that improve the source classifiers, it is important to adapt via conjugate pseudo-labeling to reap the maximum gains. 
 
 Similarly, we experiment with the case when the source classifier is trained using squared loss on the CIFAR-10 and CIFAR-100 datasets, and observe consistent gains using the proposed conjugate pseudo-labels over the baselines. For example, on CIFAR-10-C, \tta using conjugate PL gives and error of $12.87\%$, outperforming baselines like ENT ($13.24\%$) and Softmax PL ($31.81\%$). \autoref{tab:squaredloss_results} in Appendix \ref{app:square_loss} shows the detailed results.
 
 Comparing Table~\ref{tab:ce_loss} and Table~\ref{tab:polyloss1}, we see that the relative ordering between the various baselines differs. This is further evidence that the adaptation loss has to depend on the training loss, and we believe our conjugate pseudo-label approach captures this appropriately by offering consistent gains across the various settings we experimented with. 

%% file: tables/table-ce.tex
\begin{table}[h]
    \centering
    \resizebox{0.93\textwidth}{!}{
\begin{tabular}{ccccc||cc}
    \toprule
    Dataset & \multicolumn{1}{c}{\begin{tabular}[c]{@{}c@{}}Temperature \\ (T)\end{tabular}}  & \multicolumn{1}{c}{Hard PL} & \multicolumn{1}{c}{Robust PL} & \multicolumn{1}{c||}{MEMO}  & \multicolumn{1}{c}{\begin{tabular}[c]{@{}c@{}}Conjugate PL\\ (\textbf{ENT})\end{tabular}}\\
    \midrule
    \multirow{2}{*}{CIFAR-10-C} &  \xmark & 13.95 ($\pm$0.06) & 13.97 ($\pm$0.04) & \textbf{12.60} ($\pm$0.04) & 13.07 ($\pm$0.05) & \\
    & \cmark  & 13.95 ($\pm$0.06) & 12.85 ($\pm$0.04) & \textbf{12.51} ($\pm$0.01) & \textbf{12.51} ($\pm$0.03) &\\
    \midrule
    \multirow{2}{*}{CIFAR-100-C} &  \xmark & 45.22 ($\pm$0.4) & 39.80 ($\pm$0.18) & \textbf{38.52} ($\pm$0.16) & 41.15 ($\pm$0.25) & \\
    & \cmark   & 45.22 ($\pm$0.4) & 36.37 ($\pm$0.10) & 37.38 ($\pm$0.06) & \textbf{36.10} ($\pm$0.07) & \\
    \midrule
    \multirow{2}{*}{ImageNet-C} &  \xmark & \textbf{45.43} ($\pm$0.05) & 45.68 ($\pm$0.01) & 48.91($\pm$0.03) & 45.82($\pm$0.01) & \\
    & \cmark   & 45.43 ($\pm$0.05) & 45.61 ($\pm$0.01) & 48.91($\pm$0.04) & \textbf{45.36}($\pm$0.01) & \\
    \bottomrule
    \end{tabular}
    }
    \vspace{1ex}
    \caption{Mean errors when adapting to corruptions using a source classifier trained via cross-entropy loss. Here, conjugate pseudo-labeling becomes softmax-entropy minimization. With the right temperature scaling, softmax-entropy minimization matches or outperforms other approaches. Prior reported gains of other methods over softmax-entropy minimization disappear when we use temperature scaling. For additional context, the source classifier errors without adaptation are: CIFAR-10-C ($29.54\%$), CIFAR-100-C ($62.26\%$), ImageNet-C ($61.89\%$)}
    \label{tab:ce_loss}
\end{table}

%% file: tables/table-polyloss1.tex
\begin{table}[!ht]
    \centering
    \resizebox{1.0\textwidth}{!}{
\begin{tabular}{ccccccc||c}
    \toprule
    Dataset & T  & \multicolumn{1}{c}{Hard PL} & \multicolumn{1}{c}{Robust PL} & \multicolumn{1}{c}{ENT}  & \multicolumn{1}{c}{MEMO} & \multicolumn{1}{c||}{Softmax PL} & \multicolumn{1}{c}{\begin{tabular}[c]{@{}c@{}}Conjugate PL\\ (\textbf{Ours})\end{tabular}}\\
    \midrule
    \multirow{2}{*}{CIFAR-10-C} &  \xmark & 13.81($\pm$0.12) & 14.23($\pm$0.02) & 13.46($\pm$0.06) & 13.23($\pm$0.07) & 14.64($\pm$0.11) & \textbf{13.02}($\pm$0.09)  \\
    & \cmark & 13.81($\pm$0.12) & 12.45($\pm$0.05) & 12.23($\pm$0.06) & 12.33($\pm$0.04) & 12.26($\pm$0.04) & \textbf{12.08}($\pm$0.05)  \\\midrule 
    \multirow{2}{*}{CIFAR-100-C} &  \xmark & 40.47($\pm$0.05) & 42.86($\pm$0.11) & 40.12($\pm$0.08)& 39.90($\pm$0.05) & 41.00($\pm$0.11) & \textbf{38.17}($\pm$0.17) \\
    & \cmark & 40.47($\pm$0.05) & 39.80($\pm$0.08) & 38.23($\pm$0.05) & 39.23($\pm$0.04) & 37.04($\pm$0.06) & \textbf{36.83}($\pm$0.08)\\\midrule
    \multirow{2}{*}{ImageNet-C} &  \xmark & 45.44($\pm$0.21)  & 46.27($\pm$0.03)  & 46.10($\pm$0.03) & 48.21($\pm$0.05)  & 44.63($\pm$0.03) & \textbf{44.01}($\pm$0.01)\\
    & \cmark & 45.44($\pm$0.21)  & 46.27($\pm$0.03)  & 45.50($\pm$0.02) & 48.21($\pm$0.04) & 44.45($\pm$0.03) & \textbf{44.01}($\pm$0.01) \\
    \bottomrule
    \end{tabular}
    }
    \vspace{1ex}
    \caption{Mean errors when adapting to corruptions using a source classifier trained via recently proposed Poly-1 Loss \cite{polyloss}. Conjugate pseudo-labeling consistently outperforms all previous approaches. For additional context, source classifier errors without adaptation : CIFAR-10-C ($30.22\%$), CIFAR-100-C ($63.91\%$) and ImageNet-C ($62.18\%$).}
    \label{tab:polyloss1}
\end{table}
% \vspace{-2em}

%% file: tables/table-polyloss2.tex
% \setlength{\tabcolsep}{6pt}
\renewcommand{\arraystretch}{1.25}
\begin{table*}[!t]%
% \begin{table*}[!ht]
% \resizebox{0.35\textwidth}{!}{
\begin{minipage}[b]{0.63\textwidth}%
% \begin{footnotesize}
\centering
% \vspace{1ex}
\resizebox{1.0\textwidth}{!}{
\begin{tabular}{cccccc||c}
    \toprule
    Dataset  &  \begin{tabular}[c]{@{}c@{}}Source\\ Error\end{tabular} & 
    \begin{tabular}[c]{@{}c@{}}Hard\\ PL\end{tabular}  & \begin{tabular}[c]{@{}c@{}}Robust\\ PL \end{tabular} & \multicolumn{1}{c}{Entropy}  &
    \begin{tabular}[c||]{@{}c@{}}Softmax\\ PL\end{tabular} &
    \begin{tabular}[c]{@{}c@{}}Conjugate PL\\ \textbf{Ours}\end{tabular} \\
    \midrule
    \multirow{1}{*}{SVHN $\rightarrow$ MNIST} & 28.33 & 20.21  & 19.73  & 14.28 & 16.54 & \textbf{10.73} \\\midrule 
    % \multirow{1}{*}{SVHN $\rightarrow$ MNIST} & 28.33 & 20.21  & 19.73  & 14.28 & 16.54 & \textbf{10.73} \\\midrule 
    \multirow{1}{*}{SVHN $\rightarrow$ USPS} & 31.58  & 23.32  & 26.12 & 23.12 & 24.07 & \textbf{21.62} \\\midrule 
    \multirow{1}{*}{SVHN $\rightarrow$ MNISTM} & 61.69  & 50.73  & 51.35 & 49.33 & 50.47 & \textbf{47.59} \\ \midrule 
    \multirow{1}{*}{ImageNet-R} & 64.19  & 58.52  & 59.46 & 58.25 & 56.62 & \textbf{55.63} \\\midrule 
    \multirow{1}{*}{VisDA-C} & 58.13  & 40.43  & 45.44 & 44.11 & 39.63 & \textbf{38.42} \\
    \bottomrule
    \end{tabular}
}
\vspace{1.0ex}
\captionof{table}{Target error when adapting models trained via polyloss on source domains across different domain adaptation benchmarks. Conjugate pseudo-labeling offers consistent and substantial gains over previous approaches across three datasets.}
\label{tab:polyloss2}
\end{minipage}
\hspace{0.4ex}
\begin{minipage}[b]{0.35\textwidth}%
\centering

\includegraphics[width=0.95\textwidth]{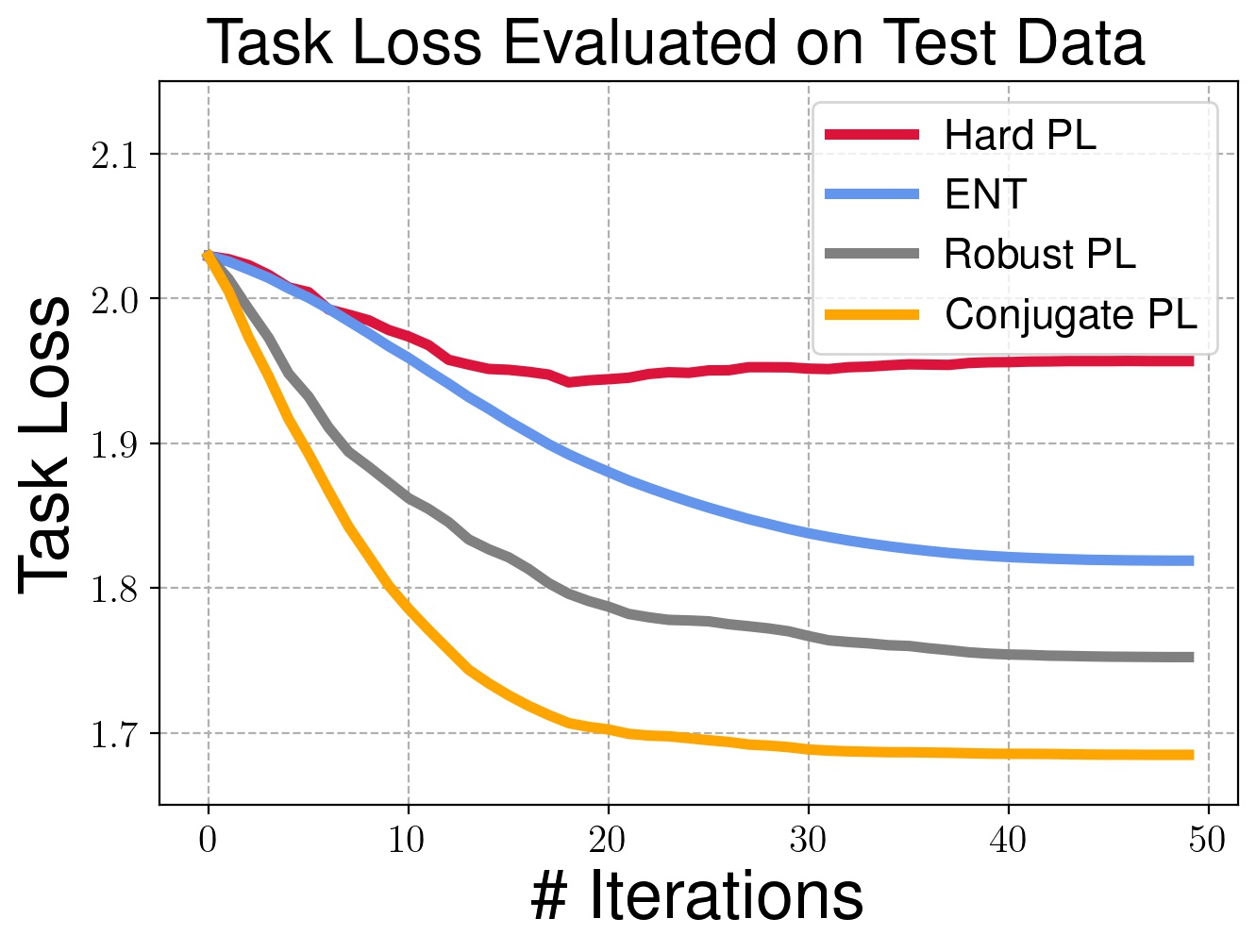}
% \vspace{3.3ex}
\captionof{figure}{Task Loss (PolyLoss $\epsilon=6$) evaluated on CIFAR-10-C test data during test-time adaptation.}
% \vspace{-0.5ex}
\label{fig:task_loss_curve}
\end{minipage}
% \vspace{-3ex}
\end{table*}

% \begin{table}[!ht]
%     \centering
%     % \setlength{\tabcolsep}{1.4pt}
%     \resizebox{0.8\textwidth}{!}{
% \begin{tabular}{cccccc||c}
%     \toprule
%     Dataset  &  \begin{tabular}[c]{@{}c@{}}Source\\ Error\end{tabular} & 
%     \begin{tabular}[c]{@{}c@{}}Hard\\ PL\end{tabular}  & \begin{tabular}[c]{@{}c@{}}Robust\\ PL \end{tabular} & \multicolumn{1}{c}{Entropy}  &
%     \begin{tabular}[c||]{@{}c@{}}Softmax\\ PL\end{tabular} &
%     \begin{tabular}[c]{@{}c@{}}Conjugate PL\\ \textbf{Ours}\end{tabular} \\
%     \midrule
%     \multirow{1}{*}{SVHN $\rightarrow$ MNIST} & 28.33 & 20.21  & 19.73  & 14.28 & 16.54 & \textbf{10.73} \\\midrule 
%     \multirow{1}{*}{ImageNet-R} & 64.19  & 58.52  & 59.46 & 58.25 & 56.62 & \textbf{55.63} \\\midrule 
%     \multirow{1}{*}{VisDA-C} & 58.13  & 40.43  & 45.44 & 44.11 & 39.63 & \textbf{38.42} \\
%     \bottomrule
%     \end{tabular}
%     }
%     \vspace{1ex}
%     \caption{Target error when adapting models trained via polyloss on source domains across different domain adaptation benchmarks. Conjugate pseudo-labeling offers consistent and substantial gains over previous approaches across three datasets. 
%     }
%     \label{tab:polyloss2}
% \end{table}
% \vspace{-1em}

%% file: related_works.tex
\section{Related Works}\label{sec:relatedwork}
% \vspace{-0.5em}
%The problem of performance drop under distribution shift has been approached under various settings with different kind of assumptions on the train/test data availability. 
\textbf{Test-time adaptation methods.} In recent years, the setting of test-time adaptation has gained a lot of interest with a host of different approaches proposed in the literature. One family of \tta approaches update the source classifier by minimizing an unsupervised loss on the target distribution~\citep{tent,rpl,memo,contrastive_tta, tta_mummadi,tta_perturb,ttpt,tta_cfa,tta_adjust,eata,continual_tta,tta_pf}.
% Although the success of \tta hinges heavily on some heuristics such as udateing ~\cite{bn_adapt,tent} (e.g., adaptation parameters), previous works are mostly focused on proposing new \textit{unsupervised} objectives. 
% \cite{ttt} optimizes over the task loss from a self-supervised learning problem, but it alters the training of source classifier. 
TENT~\citep{tent} proposes to minimize the entropy of model predictions at test time. Several follow ups like \cite{rpl, memo,contrastive_tta,tta_mummadi,tta_perturb} propose alternative \tta objectives, e.g. robust pseudo-labelling~\cite{rpl}, likelihood ratio loss~\citep{tta_mummadi}, entropy of marginal probability averaged across augmentations~\cite{memo} and self-supervised contrastive losses~\cite{on_target,contrastive_tta}. However, most of these objectives are heuristically designed or chosen. In this paper, we provide a principled approach of designing unsupervised objectives for \tta. 

Another family of approaches for test-time adaptation such as~\citep{ttt,ttt+,ttt_cs,ttt_policy,tta_video, ttt_read} leverage an auxiliary self-supervised task (e.g. rotation prediction~\citep{ttt}, masked  autoencoders~\cite{ttt_mae}) to update model parameters on each test sample. Crucially, these methods require modifying the source model training by augmenting the supervised training objective with an auxiliary self-supervised loss. Hence it cannot be applied to typical standard classifiers that are trained by minimizing a supervised loss on the source data. 

\textbf{Source-free domain adaptation.} A very related setting to test-time adaptation is source-free domain adaptation, where a trained source classifier must be adapted to a target distribution of interest, although the entire target unlabeled
data is available at once.
% ~\cite{gen1,gen2,shot} in general consider an apriori access to unlabeled data from the target domain. These approaches often require offline optimization on the whole unlabeled dataset from target distributions with carefully designed objectives and custom training pipelines. 
SHOT \cite{shot} proposes to optimize the source hypothesis (i.e. feature extractor) with a combination of entropy minimization, diversity and self-training on pseudo-labels on the unlabeled target data. \cite{clustering} promotes feature clustering on features from target distributions. \cite{gen1, gen2} use generative modeling to estimate the underlying source distributions for enforcing feature invariance. Such approaches typically require multiple epochs over the target data and cannot be easily adopted to work in an online fashion.

\textbf{Unsupervised domain adaptation.}
% For completeness, we briefly discuss distributionally robust optimization approaches. During training, if the source data has subgroups which are imbalanced, a common approach involves reweighing or sub-sampling the groups ~\citep{jtt, overparam, lip}. Model-patching ~\citep{modelpatch} learns to generate examples for minority subgroups, and enforces invariance across intra-class subgroups.  
% A related line of work focuses on various formulations of worst-case risk minimization \citep{groupDRO, drsl, class_weigh}. IRM \cite{irm} proposes to learn a feature space, such that the optimal classifier over the features is the same for all the groups. 
The most canonical setting of domain adaptation involves access to labeled source data and unlabeled target data, all during training. The availability of source and target data during training lends itself to approaches that ``align'' the source and target representations in some way: \cite{coral, mmd, mmd_2, mmd_3} match distribution statistics, \cite{dann_adv} uses a discriminator, \cite{udass} uses self-supervised learning. However, such approaches require access to source data which might not always be feasible due to data privacy and efficiency issues. 

\textbf{Pseudo-labels and self-training.}
Self-training is a classic idea for leveraging unlabeled data, developed first for the semi-supervised setting. Self-training generates pseudo-labels on the unlabeled data, allowing us to use any ``supervised'' loss on this pseudo-labeled data. Self-training has shown promising results in various settings like semi-supervised learning ~\citep{self_train_semisup} and improving adversarial robustness ~\citep{self_train_adv}. Self-training has also been gaining attention in the setting of unsupervised domain adaptation~\citep{sentry,shot}, where pseudo-labels generated on the unlabeled data from target domain is used to supervise the adaptation process. \cite{gradual_uda, self_spurious, inout} provide theoretical insights into how self-training with pseudo-labels can help under distribution shift.
TENT~\citep{tent} (i.e entropy minimization) can be viewed as a form of self-training with instantaneous softmax pseudo-labels. Our work provides a general framework for the choice of soft pseudo-labels based on the conjugate analysis of the source training objective. Some prior works like \cite{self_spurious, learning_selftrain, momentum_pl, PLO, PLT, cst} have documented the improvement in performance when using instantaneous pseudo-labels over pre-computed pseudo-labels, and thus lend further support to the benefits of our proposed conjugate pseudo-labeling approach. The experiment results presented in this work supporting conjugate pseudo-labels suggest that conjugate pseudo-labels is a promising direction of pseudo-labeling in a broader context. 

%% file: conclusion.tex
% \vspace{-0.75em}
\section{Conclusion, Limitations and Future Directions}\label{sec:limitations}

In this work, we proposed a general test-time adaptation loss, based on the convex conjugate formulation which in turn was motivated by the intriguing meta learning experiments. The fact that meta-learning recovers the proposed loss hints at some kind of optimality of the loss. In Section~\ref{sec-conjugate}, we prove that for a broad set of loss functions, the proposed (unsupervised) conjugate loss is close to the oracle supervised loss. However, this still does not completely answer what the optimal test-time adaptation loss is and why. 

The meta-learning framework in this work was constrained to learn functions over the logits of each individual input. It can be expanded to more involved setups, where we consider functions over the intermediate representations too and also consider learning functions over a batch of input while accounting for their interactions.

Beyond the choice of the adaptation loss itself, achieving good test-time adaptation generally involves several heuristics like updating only the batch norm parameters \citep{tent}. While our work was motivated by the loss function, via the meta-learning experiments, we discovered that temperature scaling is another important hyper-parameter that improves the performance of all previous baselines as well. At a high level, test-time adaptation has to be appropriately regularized to prevent the updates over batches from taking the model too far: updating only a few batch norm parameters is one way to do that, and perhaps temperature scaling provides a similar beneficial regularization effect by making the network predictions on unlabeled inputs less confident. Understanding the role of these heuristics more concretely is an interesting direction for future work. It also remains an open problem to understand under what sort of real-world distribution shifts would self-training based approaches would help.

Finally, it is also worth extending and applying the conjugate pseudo-labeling to other settings like semi-supervised learning.

\section{Acknowledgments}
We thank Shubhang Bhatnagar and Asher Trockman for helping with running the ImageNet experiments. We thank Zhili Feng for useful feedback. Sachin Goyal and Mingjie Sun were supported by funding from the Bosch Center for Artificial Intelligence. Aditi Raghunathan was supported by an Open
Philanthropy AI Fellowship.

%% file: appendix.tex
\section{Appendix}

\subsection{Conjugate Derivations}
\label{app:conjugate_derivations}
\paragraph{Cross-Entropy Loss :}

\begin{equation}
\begin{split}
\mathcal{L}(h,y) & = -\sum_{i=1}^{c}y_i\text{log}\frac{\text{exp}(h_i)}{\sum_{j=1}^{c}\text{exp}(h_j)} \\
& = -\sum_{i=1}^{c}y_i*h_i + \text{log}\sum_{j=1}^{c}\text{exp}(h_j) \\
& = f(h) - y^\top h,
\end{split}
\end{equation}

where $f(h)$ is $\text{log}\sum_{j=1}^{c}\text{exp}(h_j) $ and the constraint that $\sum_{i=1}^{c}y_i=1$. Now, the conjugate $f^\star(y)$ is given by : 
\begin{equation}
    f^\star(y) = - \min_h \{f(h) - y^T h \} =  - \min_h \{\text{log}\sum_{j=1}^{c}\text{exp}(h_j) - y^T h \}
\end{equation}
with the constraint $\sum_{i=1}^{c}y_i=1$. At the optimality, 
\begin{equation}
    y_{i} = (\nabla f(h))_i = {\text{exp}(h_{i})\over \sum_{j}\text{exp}^(h_{j})}
\end{equation}
Then, 
\begin{equation}
\begin{split}
    f^\star(y) & = - \text{log}\sum_{j=1}^{c}\text{exp}(h_j) + \sum_{i=1}^{c}h_i{\text{exp}(h_{i})\over \sum_{j}\text{exp}^(h_{j})}\\
    & = \sum_i \frac{\exp(h_i)}{\sum_j \exp(h_j)} \log \frac{\exp(h_i)}{\sum_j \exp(h_j)},
\end{split}
\end{equation}
if the constraint $\sum_{i=1}^{c}y_i=1$ is satisfied, otherwise $f^\star(y) = \infty$ by duality.

This in turn gives, the conjugate loss for cross-entropy (when the constraint is satisfied) :
\begin{equation}
    \mathcal{L}^\text{conj}(h) = -f^\star(y) = -f^\star(\nabla f(h)) = - \sum_i \frac{\exp(h_i)}{\sum_j \exp(h_j)} \log \frac{\exp(h_i)}{\sum_j \exp(h_j)}
\end{equation}

\paragraph{Squared Loss :}
\begin{equation}
\begin{split}
\mathcal{L}(h,y) & = \frac{1}{2}||h-y||_2^2 \\
& \approx \frac{1}{2}||h||_2^2 - y^\top h \text{       \hspace{2pt}[ignoring the constant term]}\\
& = f(h) - y^\top h,
\end{split}
\end{equation}
Now, the conjugate $f^\star(y)$ is given by:
\begin{equation}
\begin{split}
    f^\star(y) & = - \min_h \{f(h) - y^T h \} =  - \min_h \{\frac{1}{2}||h||_2^2 - y^T h \}\\
    & = -\frac{1}{2}||h||_2^2
\end{split}
\end{equation}

\vspace{2ex}
\subsection{Experiments on Binary Classification with Exponential Loss}
\label{app:toy_datasets}
Here we present the results on a binary classification task over a synthetic dataset of 100 dimensional gaussian clusters. 

\paragraph{Dataset Creation} \label{sec:dataset_creation}
For the binary classification task, we create a synthetic dataset similar to \citep{gradual_uda}. Specifically, let the data
$\mathcal{X}\sim \mathcal{N}(\mu,\Sigma)\in \mathbb{R}^{100}  \text{ and labels }  \mathcal{Y}\in\{-1,+1\}$. We sample $\mu\sim\mathcal{N}(k,\mathcal{I}_{100})$. For $\Sigma$, similar to \citep{gradual_uda}, we sample a diagonal matrix $D$, where each entry is sampled uniformly from a specified range, and a rotation matrix $U$ from a HAAR distribution, giving $\Sigma=UDU^T$.\\
For the source data, we sample $\mu_s^{-1}, \mu_s^{+1}, \Sigma_s^{-1}, \Sigma_s^{+1}$ as specified above with $k=0$. Now to create a distribution shifted data of various severity, we sample $\mu_t^{-1}, \mu_t^{+1}, \Sigma_t^{-1}, \Sigma_t^{+1}$ as specified above with $k=1$, which are then used to sample the shifted data as follows : 
$$\mu_{\lambda}^{1} = \lambda\mu_t^{1} + (1-\lambda)\mu_s^{1}$$
$$\mu_{\lambda}^{-1} = \lambda\mu_t^{-1} + (1-\lambda)\mu_s^{-1}$$
$$\Sigma_{\lambda}^{1} = \lambda\Sigma_t^{1} + (1-\lambda)\Sigma_s^{1}$$
$$\Sigma_{\lambda}^{-1} = \lambda\Sigma_t^{-1} + (1-\lambda)\Sigma_s^{-1}$$
$$\mathcal{X}_{\lambda}\sim \mathcal{N}(\mu_{\lambda},\Sigma_{\lambda})$$

In the following experiments, easy shift refers to $\lambda=0.6$, moderate shift to $\lambda=0.65$ and hard shift to $\lambda=0.7$.

\begin{figure}[h]
	\centering
	\begin{subfigure}[b]{1.0\textwidth}
		\includegraphics[width=0.32\textwidth]{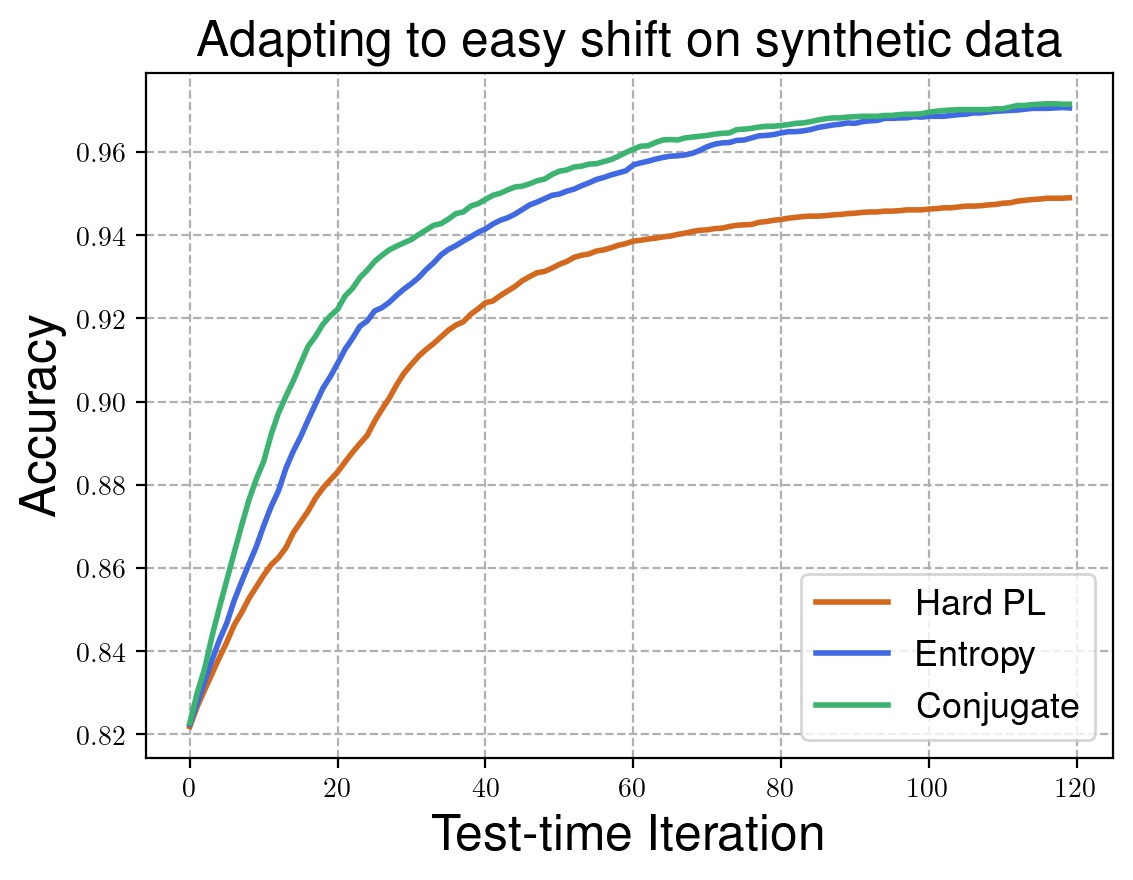}
		\includegraphics[width=0.32\textwidth]{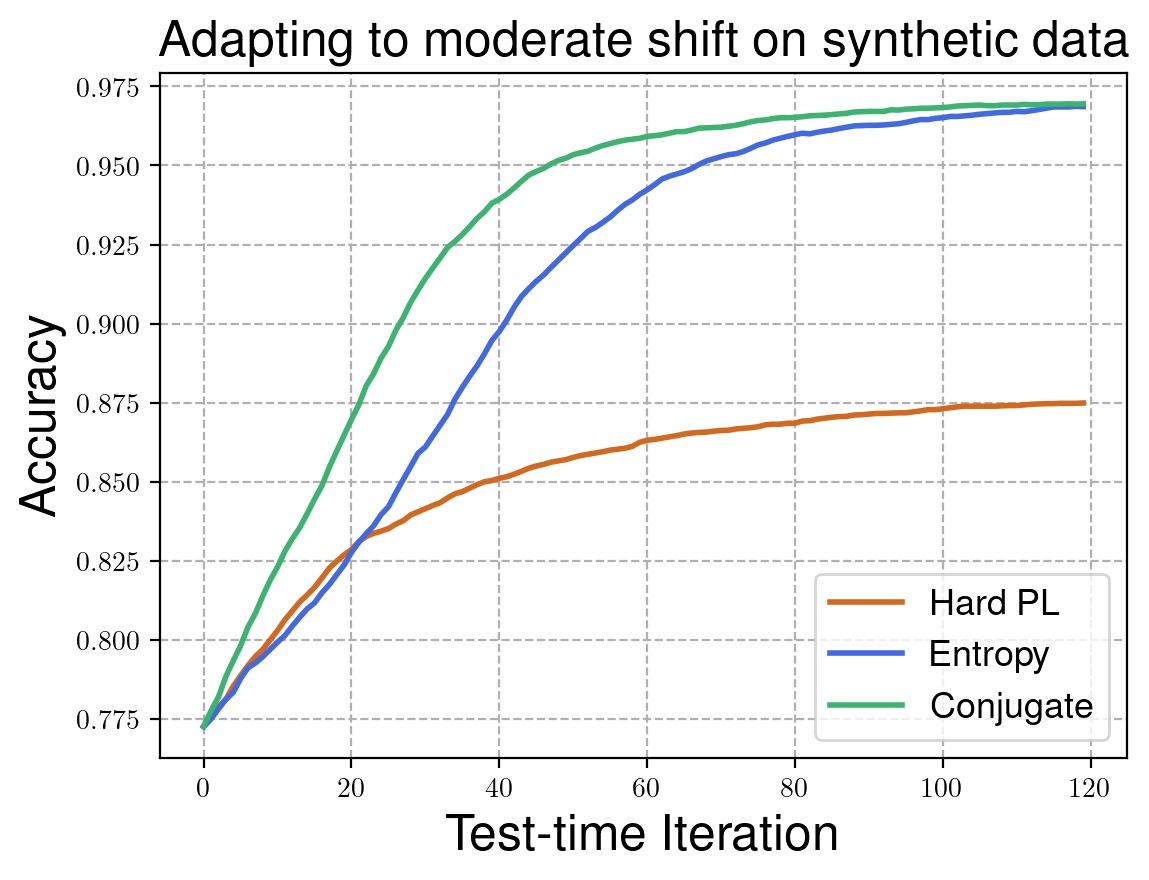}
		\includegraphics[width=0.32\textwidth]{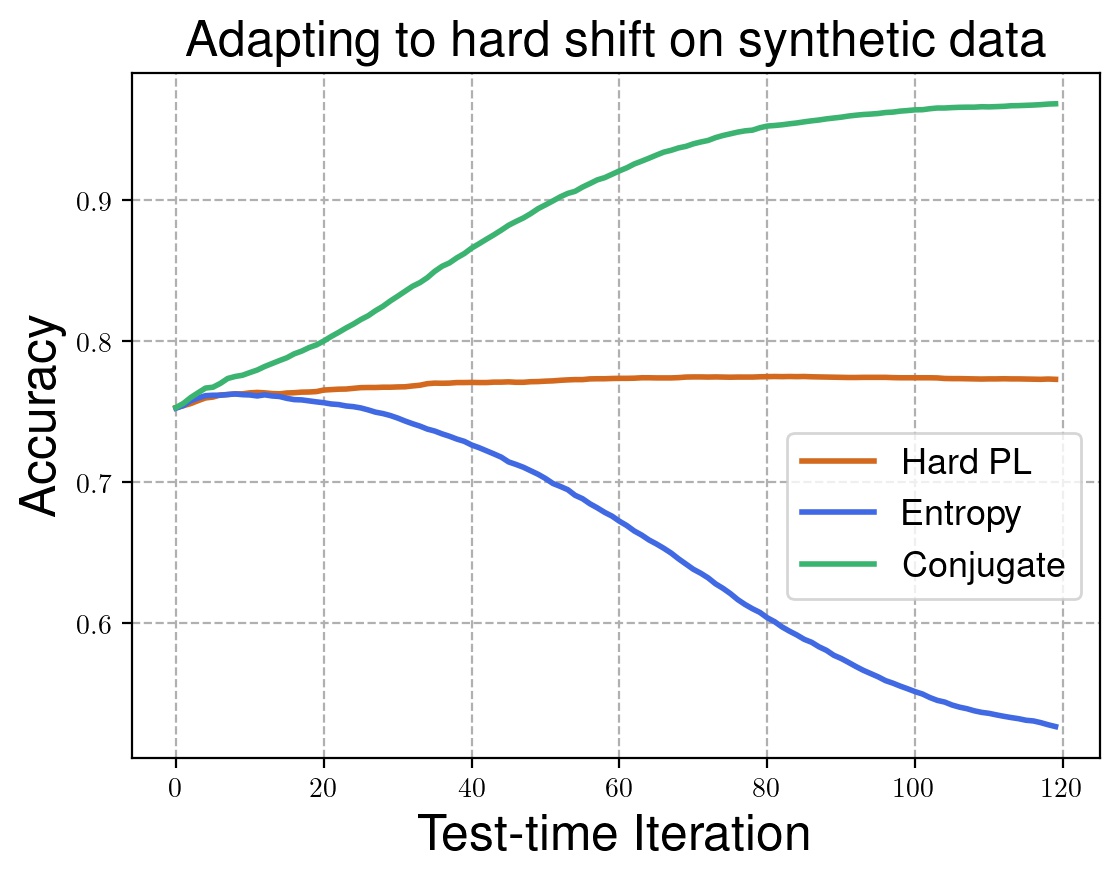}
	\end{subfigure}
  \caption{Test-time adaptation result on synthetic data with three shift levels ranging from easy, moderate and hard (detailed in \autoref{sec:dataset_creation}). The source model is a linear classifier trained with exponential loss $\mathcal{L}_\text{exp}=e^{-yh_{\theta}(x)}$. Adaptation with the conjugate loss generalizes better compared to baseline losses.}
\label{fig:gaussian_exploss}
\end{figure}

\paragraph{Exponential Loss for Binary Classification } Let $z$ be the classification score $h_{\theta}(x)$. For logistic training loss, conjugate adaptation loss would default to entropy with sigmoid probability. Thus, here we experiment with a different but also commonly used surrogate loss to 0/1 loss: exponential loss, which is defined as:  
\begin{align}
\mathcal{L}_\text{exp}(z, y)=\text{exp}(-yz)
\end{align}
where $y\in\{-1,+1\}$. It can be rewritten in the expanded conjugate form of:
\begin{align}
\mathcal{L}_\text{exp}(z,y)= \frac{1}{2}\cdot\big(e^{z}+e^{-z}\big)-\frac{1}{2}\cdot y\cdot \big(e^{z}-e^{-z}\big)
\end{align}
For exponential loss, the conjugate pseudo-label function and the conjugate pseudo-label loss are:
\begin{align}
\textstyle 
    y_\text{exp}^\text{CPL}(z) &= \frac{e^{z}-e^{-z}}{e^{z}+e^{-z}}  ,\quad \mathcal{L}_\text{exp}^\text{CPL}(z)=\frac{2}{e^{z}+e^{-z}}
\end{align}
The model is adapted on shifted gaussian clusters and we compare the conjugate loss with two  baseline approaches: 1) Hard pseudo-labelling $\text{exp}(-y_\text{hard pl}\cdot z)$; 2) Entropy applied to sigmoid probability $P(y=+1)=\sigma(z)$. The losses are compared on three degrees of shift (easy, moderate and hard), which is controlled by the drifted distance of Gaussian clusters.  
% We add an extra baseline where entropy is computed with respect to the probability $P(y=+1)=(y_\text{soft pl}+1)/2=\sigma(2z)$ corresponding to the conjugate soft pseudo label. 
% We select the hardest degree of shift, beyond which all the losses will leads to divergence. 
The results are shown in~\autoref{fig:gaussian_exploss}, where we plot the accuracy curve with respect to adaptation iterations. With easy and moderate shift, conjugate loss (green) generalizes faster to shifted test data;  with hard shift, only conjugate loss  improves model accuracy on shifted test data while entropy (blue) deteriorates model performance.

\vspace{2ex}
\subsection{Meta Learning Experiment Details }
In section \ref{sec:meta-loss} we talked about learning the meta-loss function parameterized by a neural network $m_{\phi}:\R^{|\sY|} \mapsto \R$, that takes in the model predictions/logits and outputs a loss value. Here we discuss the architecture chosen and the implementation details. Further, in Appendix \ref{sec:effect_task} we empirically show that the learnt meta-loss is not affected by the choice of task loss / surrogate loss used in meta learning ($\mathcal{L}$ in \autoref{eq:metaupdate}). Note that the task loss / surrogate loss function is used to update the meta-loss $m_{\phi}$ during meta-learning. The surrogate loss is calculated on updated source model's predictions on labeled samples from test domain. The surrogate loss tries to update the meta-loss in the outer loop such that when meta-loss is later used to update the source model in the inner loop, the source model generalizes better to the test domain.

\begin{figure}[ht]
\begin{center}
  \includegraphics[width=\textwidth]{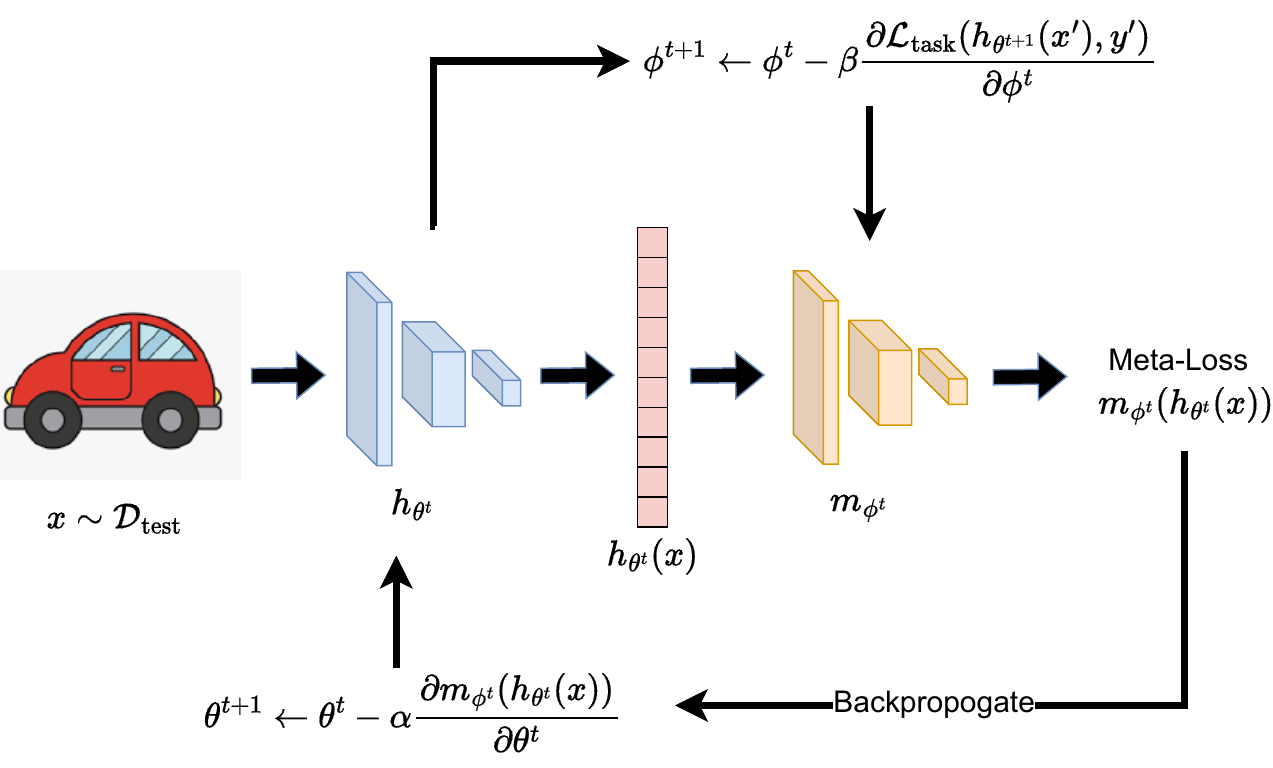}
  \caption{Meta-Loss learning procedure : The model predictions $h_{\theta^t}(x)$ are passed through the parameterized loss function $m_{\phi^t}$, which outputs a loss value. We optimize $\phi$ such that when optimizing the source model over the loss $m_{\phi^t}(h_{\theta^t}(x))$, the updated $\theta^{t+1}$ has a better performance on the test domain. To do this, we take one gradient step over the meta-loss to get the update source model parameters $\theta^{t+1}$, and then update $\phi$ by evaluating $\theta^{t+1}$ on the labeled validation data using some task loss $\mathcal{L}_\text{task}$.}
\label{fig:meta_learning}
\end{center}
\end{figure}

\paragraph{Architecture and Implementation Details } Figure \ref{fig:meta_learning} gives an overall schema for meta-learning the loss function and \autoref{alg:meta_learn} gives the pseudo-code for meta-learning the loss function. Below we describe this in further detail. We use a transformer (denoted by $\mathcal{T}$) with a MLP (denoted by $\mathcal{P}$) over the output of transformer as the architecture for $m_{\phi}$, i.e.  $m_{\phi}(x) = \mathcal{P} (\mathcal{T}(x))$. Specifically, for a given source trained model $h_\theta$ and input $x \sim \dtest$ :
\begin{enumerate}[leftmargin=*]
    \item Let $h_{\theta}(x)\in\mathbb{R}^{|\sY|}$ be the model predictions/logits, where $|\sY|$ denotes the number of classes.
    \item Let $h_{\theta}^j(x) \in \mathbb{R}, \forall j\in|\sY| $ be the prediction corresponding to class $j$.
    \item The input to transformer is then given by $z\in \mathbb{R}^{|\sY|\times (1+e)}$, where $z^j\in \mathbb{R}^{1+e}, \forall j\in|\sY|$ is the concatenation of $h_{\theta}^j(x)$ and the learnable positional embedding $pe_j\in \mathbb{R}^e$.
    \item The transformer output is given by $w = \mathcal{T}(z)\in \mathbb{R}^d$, where $d$ denotes the feed-forward dimension of the transformer.
    \item The transformer output $w$ is finally passed through a MLP to get the meta-loss value $m_{\phi}(h_{\theta}(x)) = \mathcal{P}(w) \in \mathbb{R}$
    \item The source model is updated by optimizing over the meta-loss.
    \begin{align}
    \theta^{t+1} \leftarrow\theta^t-\alpha\frac{\partial m_{\phi^t}(h_{\theta^{t}}(x))}{\partial \theta^t} 
    \end{align}
    \item The updated source model is then used to update the meta-loss by optimizing over some supervised loss function $\mathcal{L}_\text{task}$. 
    \begin{align}
    \phi^{t+1} \leftarrow \phi^t - \beta\frac{\partial \mathcal{L}_\text{task}(h_{\theta^{t+1}}(x'), y')}{\partial \phi^t},~~ \text{where } (x',y')\sim \dtest
    \end{align}
\end{enumerate}

Note that the last step assumes access to labels of test inputs. In this paper, we do not propose meta-learning the TTA loss as an approach. Rather, we use meta-learning to explore what the “best” TTA losses
look like.

We select the trasformer input embedding dimension $(1+e)$ from $\{16,32,64\}$ and transformer feed-forward dimension $d$ from $\{32,64,128\}$. The number of transformer layers and the hidden layers in MLP are selected from $\{1, 2\}$. We use Adam optimizer with a learning rate of $1e^{-3}$ for learning the meta-loss (i.e. the transformer + MLP). We train the meta-loss for 100 epochs with a batch size of 200.  

\begin{algorithm}
  \caption{Learning the Meta-Loss}
  \label{alg:meta_learn}
    \hspace*{\algorithmicindent}
    \textbf{Input:} Source trained classifier $h_{\theta^0}$. Randomly initialized meta-loss $m_{\phi^0}$.\\ 
    \hspace*{\algorithmicindent} \hspace*{\algorithmicindent} \hspace*{\algorithmicindent} Task loss / Surrogate loss $\mathcal{L}_\text{task}$ like cross-entropy or squared loss for meta learning\\
    \hspace*{\algorithmicindent} \hspace*{\algorithmicindent} \hspace*{\algorithmicindent} $N$ batches of test data $\dtest = [(x_1,y_1), \hdots, (x_N, y_N)]$
    \\
    \hspace*{\algorithmicindent} \textbf{Hyperparams:}
    learning rates $\alpha$ and $\beta$.
  \begin{algorithmic}[1]
    % \STATEx 
    % Let $\scaledpred_\theta(x) \eqdef h_\theta(x)/T$ be the temperature scaled predictor. 
    % \STATEx
    % \vspace{2pt}
    % Let $\cpl_\theta(x)$ denote the conjugate pseudo-label function $\cpl_\theta(x) =\nabla(f(\bar{h}_\theta(x)))$.
    \STATEx
    \vspace{2pt}
    \textbf{for} $epoch = 0, 1, 2, \hdots$ \textbf{do}
    \STATEx
    \vspace{2pt}
    \hspace{10pt}\textbf{for} $n = 0, 1, \hdots N-1$ \textbf{do}
    \STATEx
    \vspace{4pt}
    \hspace{20pt} $\theta^{t+1} \leftarrow\theta^t-\alpha\frac{\partial m_{\phi^t}(h_{\theta^{t}}(x_n))}{\partial \theta^t}$
    \vspace{4pt}
    \STATEx
    \hspace{20pt} Sample $(x_r,y_r) \sim \dtest$.
    \STATEx
    \vspace{4pt}
    \hspace{20pt} $\phi^{t+1} \leftarrow \phi^t - \beta\frac{\partial \mathcal{L}_\text{task}(h_{\theta^{t+1}}(x_r), y_r)}{\partial \phi^t}$
    \vspace{2pt}
  \end{algorithmic}
\end{algorithm}

\subsection{Effect of Task Loss in Meta Learning}
\label{sec:effect_task}
In ~\autoref{sec:meta-loss}, we show that the meta losses learned on different source classifiers differ substantially if the source classifiers are trained using different source loss functions. Here we further empirically verify that the learnt meta loss is not affected by the task loss used in meta learning ($\mathcal{L}$ in \autoref{eq:metaupdate}). Thus the learnt meta loss is  determined by the source model.

\begin{figure}[!htbp]
	\centering
	\begin{subfigure}{0.45\textwidth}
		\includegraphics[width=\textwidth]{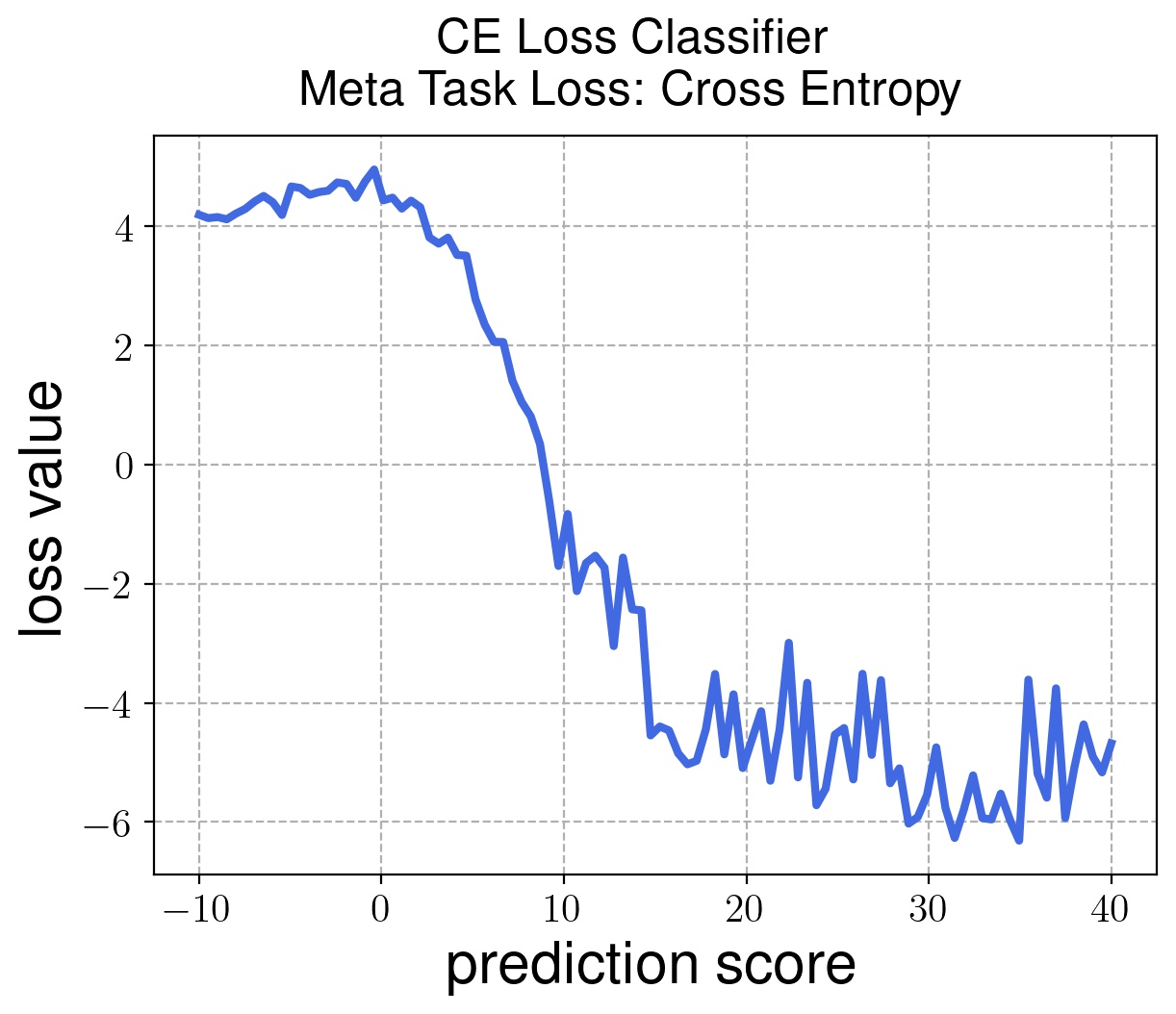}
		\caption{}
		\label{fig-appendix:meta-loss-ce-task-loss-ce}
	\end{subfigure}
%%%%%%%%%%%%%%
	\begin{subfigure}{0.45\textwidth}
		\includegraphics[width=\textwidth]{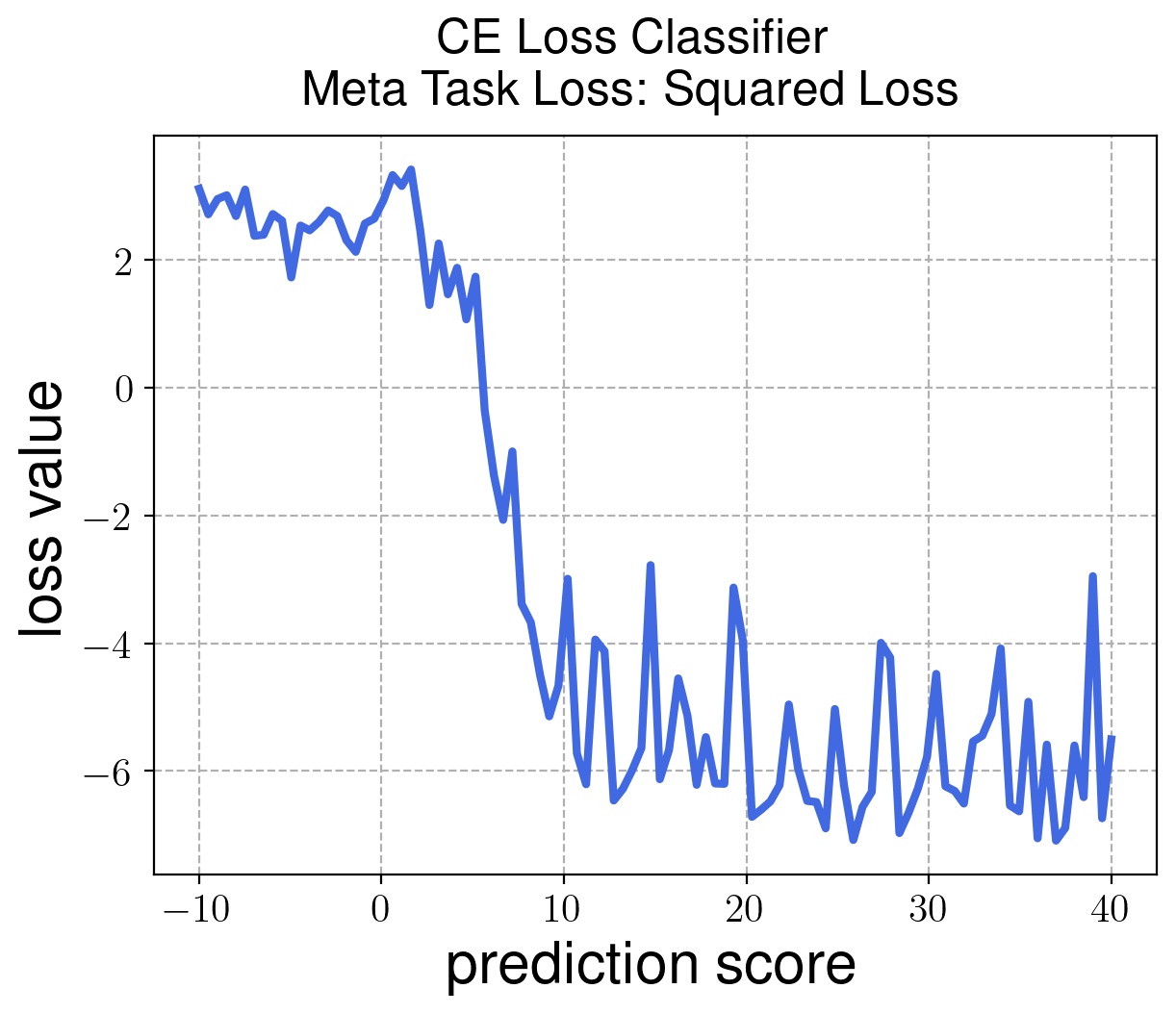}
		\caption{}
		\label{fig-appendix:meta-loss-ce-task-loss-quad}
	\end{subfigure}
	\vspace{-1ex}
  \caption{Visualizations of meta loss by varying one input dimension (prediction score). The source model is a ResNet-26 trained with Cross Entropy. Here we show meta loss trained by two different task losses: Cross Entropy~\autoref{fig-appendix:meta-loss-ce-task-loss-ce} and Squared Loss~\autoref{fig-appendix:meta-loss-ce-task-loss-quad}.}
\label{fig-appendix:meta-loss-ce}
\end{figure}

\begin{figure}[!htbp]
	\centering
	\begin{subfigure}{0.45\textwidth}
		\includegraphics[width=\textwidth]{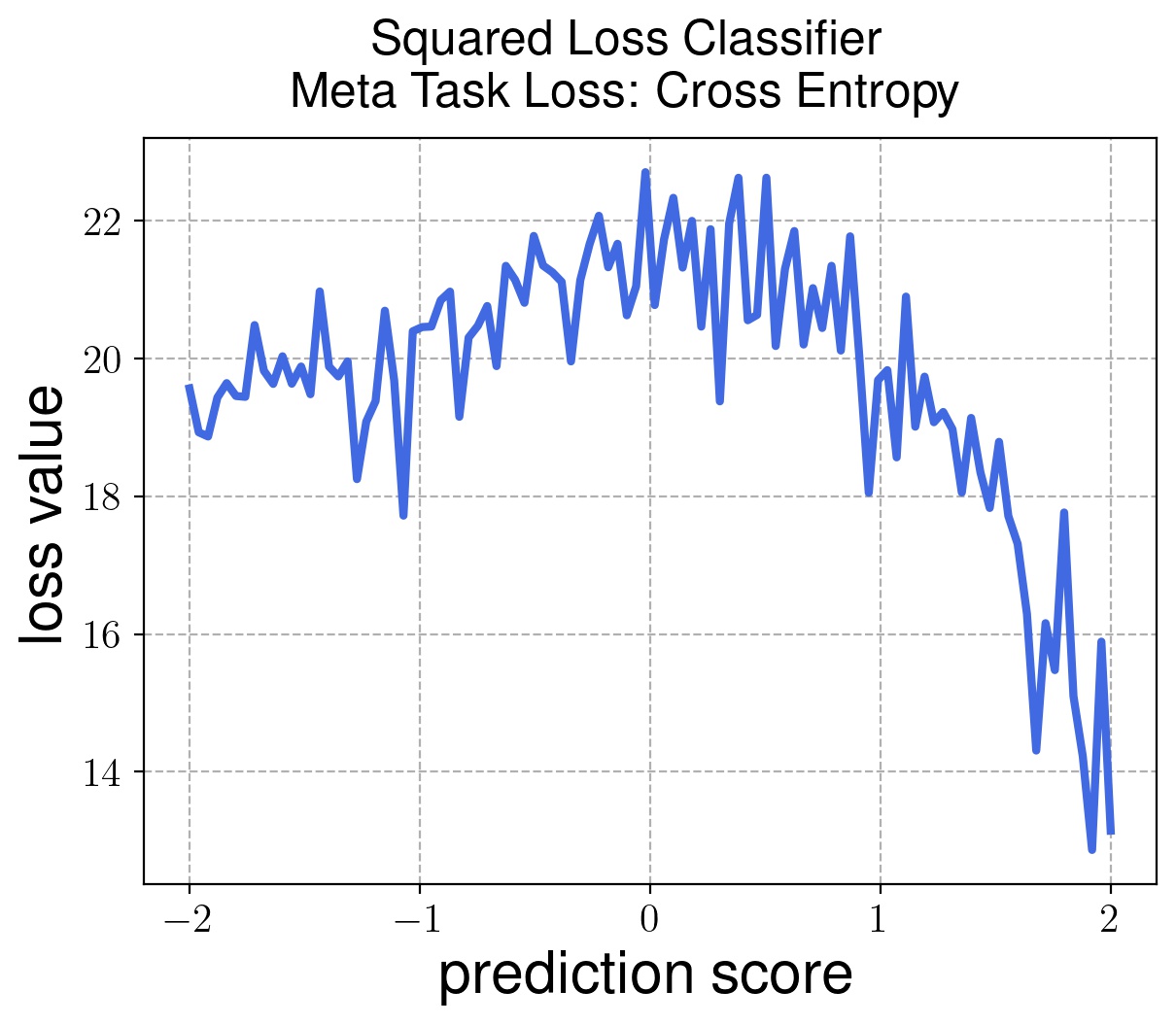}
% 		\vspace{-1ex}
		\caption{}
		\label{fig-appendix:meta-loss-quad-task-loss-ce}
	\end{subfigure}
%%%%%%%%%%%%%%
	\begin{subfigure}{0.45\textwidth}
		\includegraphics[width=\textwidth]{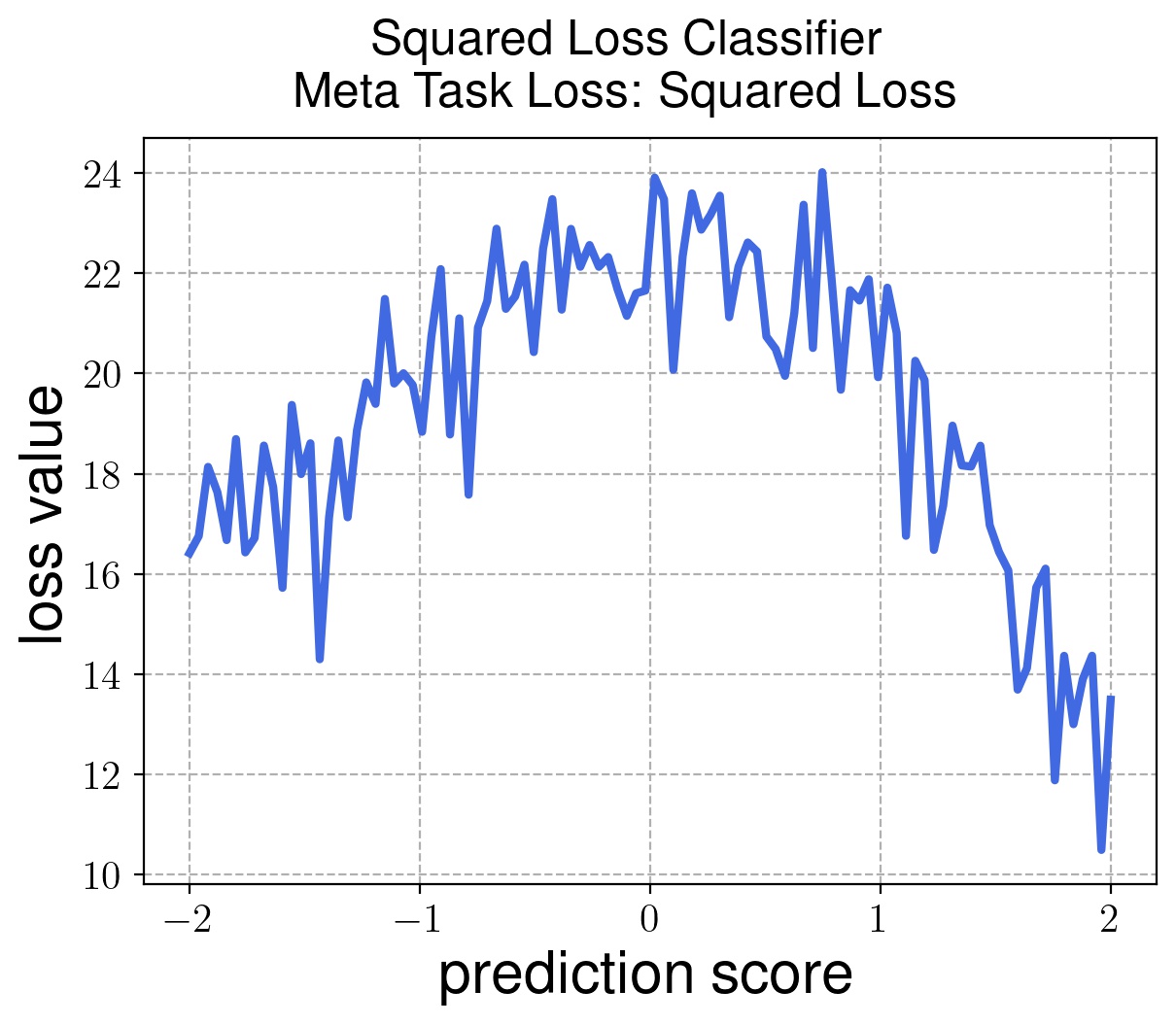}
% 		\vspace{-1ex}
		\caption{}
		\label{fig-appendix:meta-loss-quad-task-loss-quad}
	\end{subfigure}
	\vspace{-1ex}
  \caption{Visualizations of meta loss by varying one input dimension (prediction score). The source model is a ResNet-26 trained with Squared Loss. Here we show meta loss trained by two different task losses: Cross Entropy~\autoref{fig-appendix:meta-loss-quad-task-loss-ce} and Squared Loss~\autoref{fig-appendix:meta-loss-quad-task-loss-quad}.}
\label{fig-appendix:meta-loss-quad}
\end{figure}

In ~\autoref{fig-appendix:meta-loss-ce}, we show the meta loss learnt on a ResNet-26 trained with Cross Entropy loss for two meta task losses: Cross Entropy~\autoref{fig-appendix:meta-loss-ce-task-loss-ce} and Squared Loss~\autoref{fig-appendix:meta-loss-ce-task-loss-quad}. We plot the meta loss as a function over one of its input prediction scores, while keeping other fixed. We can see that the task loss barely affects the learnt meta loss. Similar observations can be made for the classifier trained with squared loss~\autoref{fig-appendix:meta-loss-quad}.

\subsection{Test-Time Adaptation Detail}
For completeness, we also give the test-time adaptation setup in Algorithm~\ref{alg:tta_alg}.

\begin{algorithm}
  \caption{Test-Time Adaptation}
  \label{alg:tta_alg}
    \hspace*{\algorithmicindent}\textbf{Input:} Source classifier $\sourcetheta$ trained using loss $\mathcal{L}(h_\theta(x), y) $, An unsupervised loss function for \hspace*{\algorithmicindent} test-time adaptation $\mathcal{L}_\text{tta}(x)$, $N$ batches of test data $\dtest = [x_1, \hdots, x_N]$
    \\
    \hspace*{\algorithmicindent} \textbf{Hyperparams:}
    learning rate $\eta$.
  \begin{algorithmic}[1]
    \STATEx
    \vspace{2pt}
    \textbf{for} $n = 0, 1, \hdots N-1$ \textbf{do}
    \STATEx
    \vspace{2pt}
    \hspace{10pt} $\theta_{n+1} = \theta_n - \eta \nabla \mathcal{L}_\text{tta}(x_n)$
    \STATEx
    \hspace{10pt}
    \vspace{-1ex}
    \STATEx
    \hspace{10pt} $\hat{y}_n = h_{\theta_{n+1}}(x_n)$ ~~\text{[Predictions for the $n^{th}$ batch]}
  \end{algorithmic}
\end{algorithm}

\subsection{ImageNet results on each severity level}
In continuation with results shown in  \autoref{tab:polyloss1} in Section 5.3,  \autoref{tab:poly1_imgnet_allsev} shows the mean errors averaged across the 15 corruption types for each of the severity level on ImageNet-C, for a source classifier trained with PolyLoss ($\epsilon=8$).

\begin{table}[!ht]
    \centering
    \resizebox{0.8\textwidth}{!}{
    \begin{tabular}{ccccccc}
    \toprule
\begin{tabular}[c]{@{}l@{}}Corrution \\ Severity\end{tabular}  & Temperature & \multicolumn{1}{c}{Robust PL} & \multicolumn{1}{c}{Entropy}  & \multicolumn{1}{c}{MEMO} & \multicolumn{1}{c}{Softmax PL} & \multicolumn{1}{c}{Conjugate}\\
\midrule
        \multirow{2}{*}{1} &  \xmark & 34.27 & 33.17 & 34.39 & 32.49 & \textbf{32.26} \\
    & \cmark & 34.27 & 32.84& 34.39 & 32.70 & \textbf{32.26} \\\midrule 
    \multirow{2}{*}{2} &  \xmark & 41.25 & 39.04 & 40.38 & 37.78 & \textbf{37.40} \\
    & \cmark & 41.25 & 38.50 & 40.38 & 37.75 & \textbf{37.40} \\\midrule   
    \multirow{2}{*}{3} &  \xmark & 47.37 & 44.04 & 45.67 & 42.30 & \textbf{41.72} \\
    & \cmark & 47.37 & 43.33& 45.67 & 42.14 & \textbf{41.72} \\\midrule 
    \multirow{2}{*}{4} &  \xmark & 56.63 & 51.88 & 54.49 & 49.61 & \textbf{48.84} \\
    & \cmark & 56.63 & 51.03 & 54.49 & 49.39 & \textbf{48.84} \\\midrule 
    \multirow{2}{*}{5} &  \xmark & 67.11 & 62.53 & 66.13 & 60.94 & \textbf{59.90} \\
    & \cmark & 67.11 & 61.80 & 66.13 & 60.30 & \textbf{59.90} \\\midrule 
    \multirow{2}{*}{Mean} &  \xmark & 49.32 & 46.13 & 48.21 & 44.62 & \textbf{44.02} \\
    & \cmark & 49.32 & 45.50 & 48.21 & 44.45 & \textbf{44.02} \\
        \bottomrule
    \end{tabular}
    }
    \vspace{1ex}
    \caption{Mean Errors across the 15 noises for various severity level on the ImageNet-C dataset, with source model trained using Poly-1 Loss. Note that Temperature scaling helped only in the case of Entropy and Softmax PL.}
    \label{tab:poly1_imgnet_allsev}
\end{table}

\subsection{Square Loss Trained Source Classifier}
\label{app:square_loss}

In Section 5.3, we briefly discussed that similar to the other source training losses like cross-entropy and polyloss, our proposed conjugate loss outperforms the baselines when the source classifier is trained using a squared loss. \autoref{tab:squaredloss_results} shows a detailed comparison with the baselines. We note that for the conjugate of squared loss, the temperature scaling can be wrapped into the learning rate as shown in Section \ref{sec:cpl_sqloss}. Further, on the CIFAR-10-C dataset we observe temperature scaling doesn't help any of the other baselines too, hence we do not include the temperature row in CIFAR-10-C.

\begin{table}[!ht]
    \centering
    \setlength{\tabcolsep}{1.4pt}
    \resizebox{1.00\textwidth}{!}{
\begin{tabular}{cccccccc}
    \toprule
    Dataset & Temperature  &  \multicolumn{1}{c}{Hard PL}  & \multicolumn{1}{c}{Robust PL} & \multicolumn{1}{c}{ENT}  & \multicolumn{1}{c}{MEMO} &
    \multicolumn{1}{c}{Softmax PL} &
    \multicolumn{1}{c}{Conjugate PL}\\
    \midrule
    \multirow{1}{*}{CIFAR-10-C} &  \xmark & 13.71 ($\pm$0.07) & 13.06 ($\pm$0.05) & 13.24 ($\pm$0.02) & 13.22 ($\pm$0.04) & 14.85 ($\pm$0.08) &  \textbf{12.99} ($\pm$0.04)  \\\midrule 
    \multirow{2}{*}{CIFAR-100-C} &  \xmark & 50.82 ($\pm$0.31) & 44.53 ($\pm$0.13) & 43.55 ($\pm$0.12)& 51.35 ($\pm$0.04) & 51.99 ($\pm$0.03) & \textbf{43.39} ($\pm$0.11) \\
    & \cmark & 50.82 ($\pm$0.31) & 43.99 ($\pm$0.15) & \textbf{43.21} ($\pm$0.08)& 51.35 ($\pm$0.04) & 51.99 ($\pm$0.03) & 43.39 ($\pm$0.11)  \\
    
    \bottomrule
    \end{tabular}
    }
    \vspace{1ex}
    \caption{Mean Errors on the common corruptions datasets for source classifier trained using squared loss. We note that temperature scaling didn't help on the CIFAR-10-C dataset. Source Classifier Errors without adaptation : CIFAR-10-C ($28.34\%$), CIFAR-100-C ($68.79\%$)}
    \label{tab:squaredloss_results}
\end{table}

\input{tables/table-ce-hparams}

\vspace{2ex}
\subsection{Hyper-Parameters}
We share the exact hyper-parameters found using gridsearch over the 4 validation noises for the common corruptions dataset. 

\paragraph{Cross Entropy Classifier Experiments} In Section 5.2, \autoref{tab:ce_loss} shows the results when adapting a cross entropy trained classifier on various common corruptions dataset. \autoref{tab:ce_loss_hparams} gives the optimizer, learning rate and optimal temperature for each of the baseline and our proposed conjugate loss.

\paragraph{PolyLoss Classifier Experiments} In Section 5.3, \autoref{tab:polyloss1} shows the results when adapting a polyloss trained classifier on various common corruptions dataset. \autoref{tab:poly_loss_hparams} gives the optimizer, learning rate and optimal temperature for each of the baseline and our proposed conjugate loss.

\input{tables/table-polyloss1-hparams}

\vspace{4ex}
\paragraph{Squared Loss Classifier Experiments} In Section 5.3, we briefly discussed the results when adapting a squared loss trained classifier on various common corruptions dataset. \autoref{tab:squared_loss_hparams} gives the optimizer, learning rate and optimal temperature for each of the baseline and our proposed conjugate loss for the results in \autoref{tab:squaredloss_results}.

\input{tables/table-quad-hparams}

\vspace{4ex}
\paragraph{Digit Adaptation Datasets} For the experiments on digits adaptation tasks, we do not have any validation set. Hence, we don't use temperature scaling here ($T=1$) and fix the optimizer and LR as Adam and $1e^{-2}$ respectively for all the baselines.

\subsection{Additional Experiments on Digit Adaptation Datasets}
\label{app:additional_digit_adapt}
Similar to the setting of Table \ref{tab:ce_loss}, we perform additional experiments on digit adaptation datasets when the source classifier is trained using the cross-entropy loss. Note that when the source classifier is trained using cross-entropy loss, the conjugate loss is equal to the softmax-entropy. In the absence of validation dataset in digit adaptation benchmarks, we used a fixed learning rate of 0.01 for all the baselines, optimizer as Adam and an informed temperature scaling guess of T=2. 

Table \ref{tab:ce_loss_digit} compares softmax-entropy minimization with various baselines. Here, again we observe that on SVHN $\rightarrow$ MNIST benchmark, without temperature scaling, MEMO ($10.67\%$ error) outperforms softmax-entropy ($14.41\%$ error). However, similar to the observations in Table \ref{tab:ce_loss}, with temperature scaling, softmax-entropy minimization ($9.26\%$ error) is able to match the performance of MEMO ($9.36\%$ error). Further, on the SVHN $\rightarrow$ USPS benchmark, softmax-entropy (conjugate) and MEMO perform similar even without temperature scaling. 

\input{tables/table-ce-digit}

\begin{figure}[!htbp]
	\centering
	\begin{subfigure}{0.45\textwidth}
		\includegraphics[width=\textwidth]{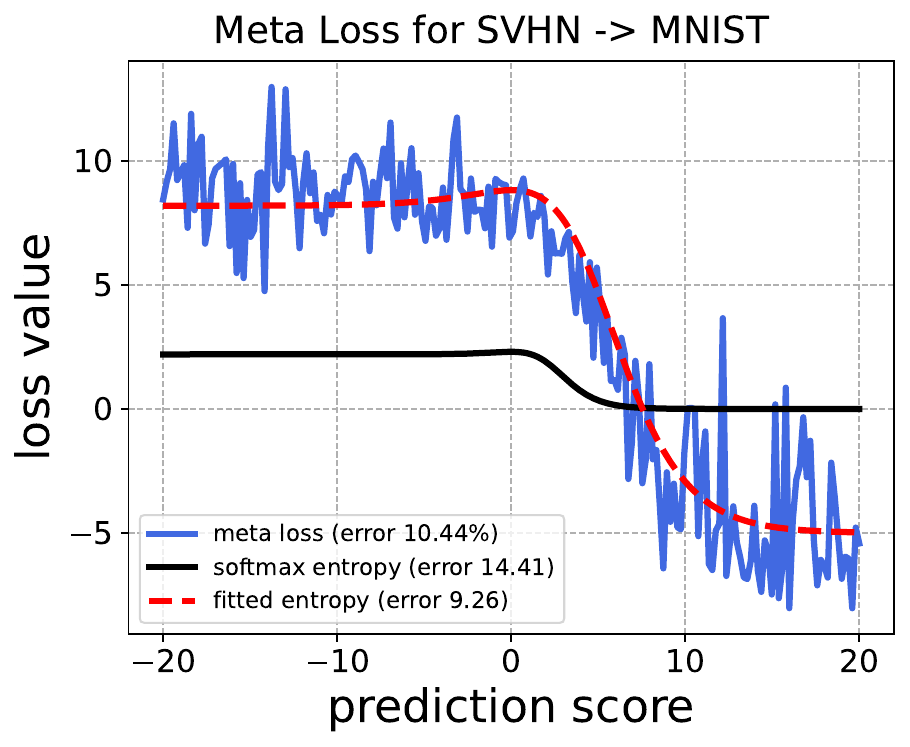}
		\caption{}
		\label{fig-appendix:svhn_mnist}
	\end{subfigure}
%%%%%%%%%%%%%%
	\begin{subfigure}{0.45\textwidth}
		\includegraphics[width=\textwidth]{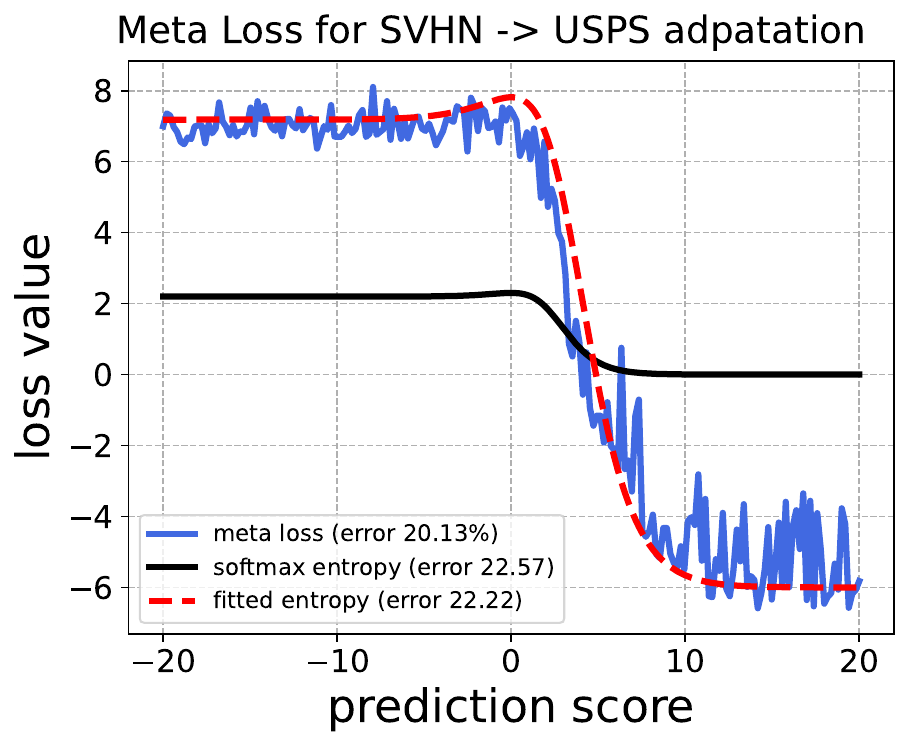}
		\caption{}
		\label{fig-appendix:svhn_usps}
	\end{subfigure}
	\vspace{-1ex}
  \caption{Visualizations of the learnt meta-loss by varying one input dimension (prediction score). The source model is a ResNet-26 trained with cross-entropy on the SVHN dataset. (a) The learnt meta-loss when adapting to the MNIST test dataset. (b) The learnt meta-loss when adapting to the USPS test dataset.}
\label{fig-appendix:svhn_adapt}
\end{figure}

\subsection{Additional Meta Learning the TTA Loss Experiments}
\label{app:additional_meta_learn}
In Section \ref{sec:meta-loss}, we tried to learn a test-time adaptation (TTA) loss via meta-learning for adapting a CIFAR10 trained ResNet26 to distribution shifts on CIFAR10 corruptions. Figure \ref{fig:fitted_ce} showed that the learnt meta-loss looks like a temperature scaled softmax-entropy.

In this section, we show the learnt meta loss across a range of settings as described below : 

\begin{enumerate}[leftmargin=*]

    \item Digit Adaptation: Figure \ref{fig-appendix:svhn_mnist} and \ref{fig-appendix:svhn_usps} show the learnt meta-loss when adapting a SVHN trained ResNet26 to MNIST dataset and USPS dataset respectively. We observe that the learnt meta-loss can be well approximated by a temperature scaled softmax-entropy. 
    
    \item Various Noise Types: In ~\autoref{appendix:resnet26_meta_loss_noise}, we show the learnt meta-loss when adapting a ResNet26 trained on CIFAR10 dataset using cross-entropy loss, to various noise types like speckle, gaussian, saturate and spatter. The severity level is kept fixed at the maximum i.e. 5.
    
    \item Various Severity Levels: In ~\autoref{appendix:resnet26_meta_loss_sev}, we vary the severity level of the noise, keeping the noise type fixed. 
    
    \item Dataset and Architecture: In ~\autoref{appendix:resnet26_meta_loss_dataset_arch}, we compare the learnt meta-loss when adapting to speckle noise, for different source classifier architectures (ResNet26 and ResNet50) and different source training dataset (CIFAR10 and CIFAR100). In all the cases, we again observe that the learnt meta-loss can be well approximated by a temperature scaled softmax-entropy.
    
    \item Squared Loss : Finally, in ~\autoref{appendix:resnet26_meta_loss_quad} we show the learnt meta-loss for classifiers trained with squared loss function instead of cross-entropy. We observe that in this case, the learnt meta loss mimics a quadratic function as expected from the conjugate formulation.
    
\end{enumerate}

 For each of the learnt meta losses, we also show the values ($\alpha, T, C$) we use to fit the meta loss with softmax entropy function: $\alpha\cdot \mathcal{H}(\text{softmax}(x/T))-C$. Note that although the learnt meta-loss can be approximated by the conjugate, the parameters $\alpha, T, C$ differ across the settings.

 In the case of classifiers trained with squared loss, we fit the meta loss with a quadratic function $\sum_{i=1}^{K} (A\cdot x_{i}^{2} + C)$, where $K$ is the number of classes and $x$ is the logit vector. Again, we also show the fitted parameter value $A, C$. The meta loss follows the trend of a quadratic function. The fitted quadratic function performs better or similar as the meta loss, while the parameters of the fitted quadratic function remain different across the meta learning setup (base classifier architectures and noise types).

\begin{figure}[!htbp]
	\centering
	\begin{subfigure}{0.40\textwidth}
		\includegraphics[width=\textwidth]{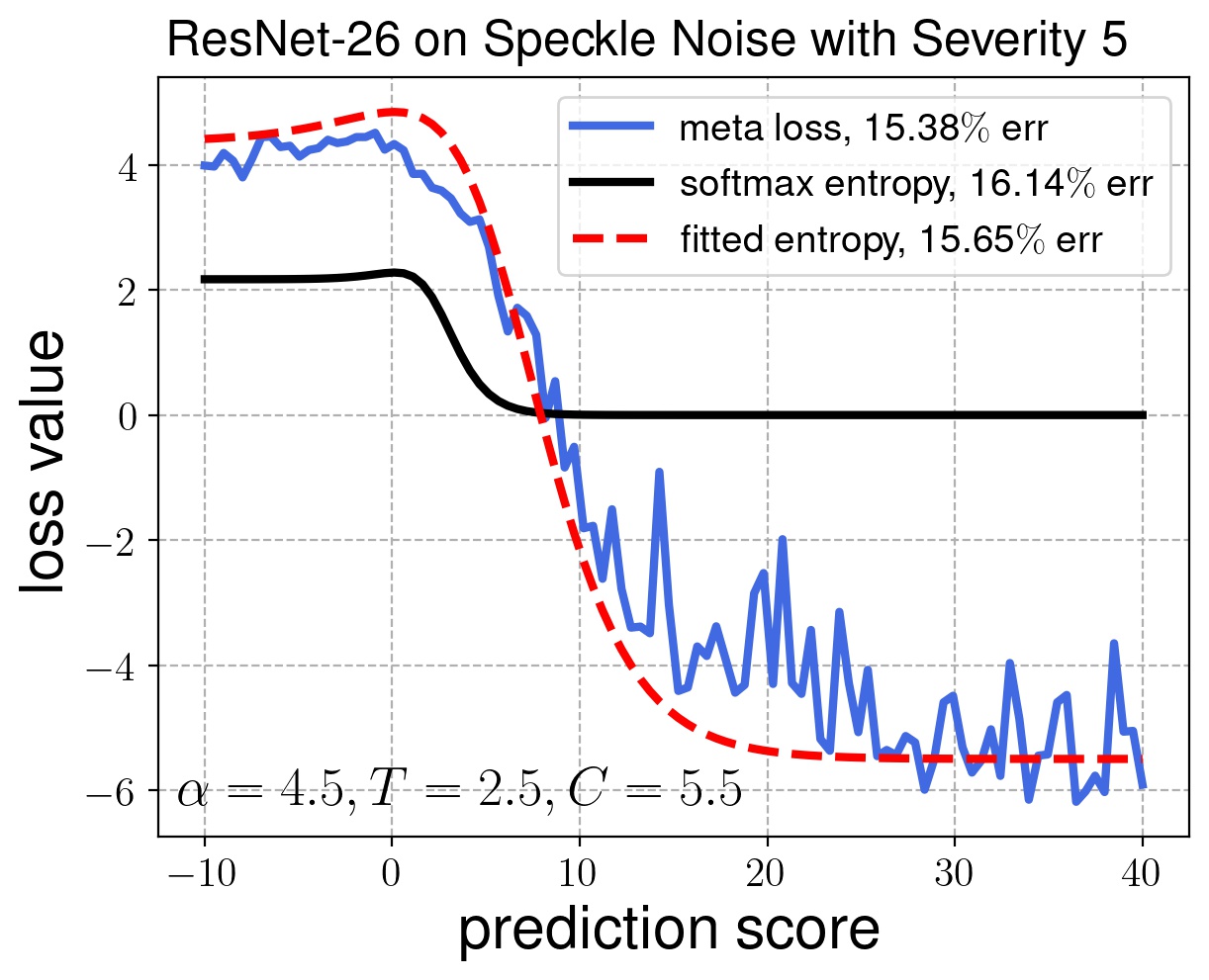}
% 		\vspace{-1ex}
		\caption{}
	\end{subfigure}
%%%%%%%%%%%%%%
	\begin{subfigure}{0.40\textwidth}
		\includegraphics[width=\textwidth]{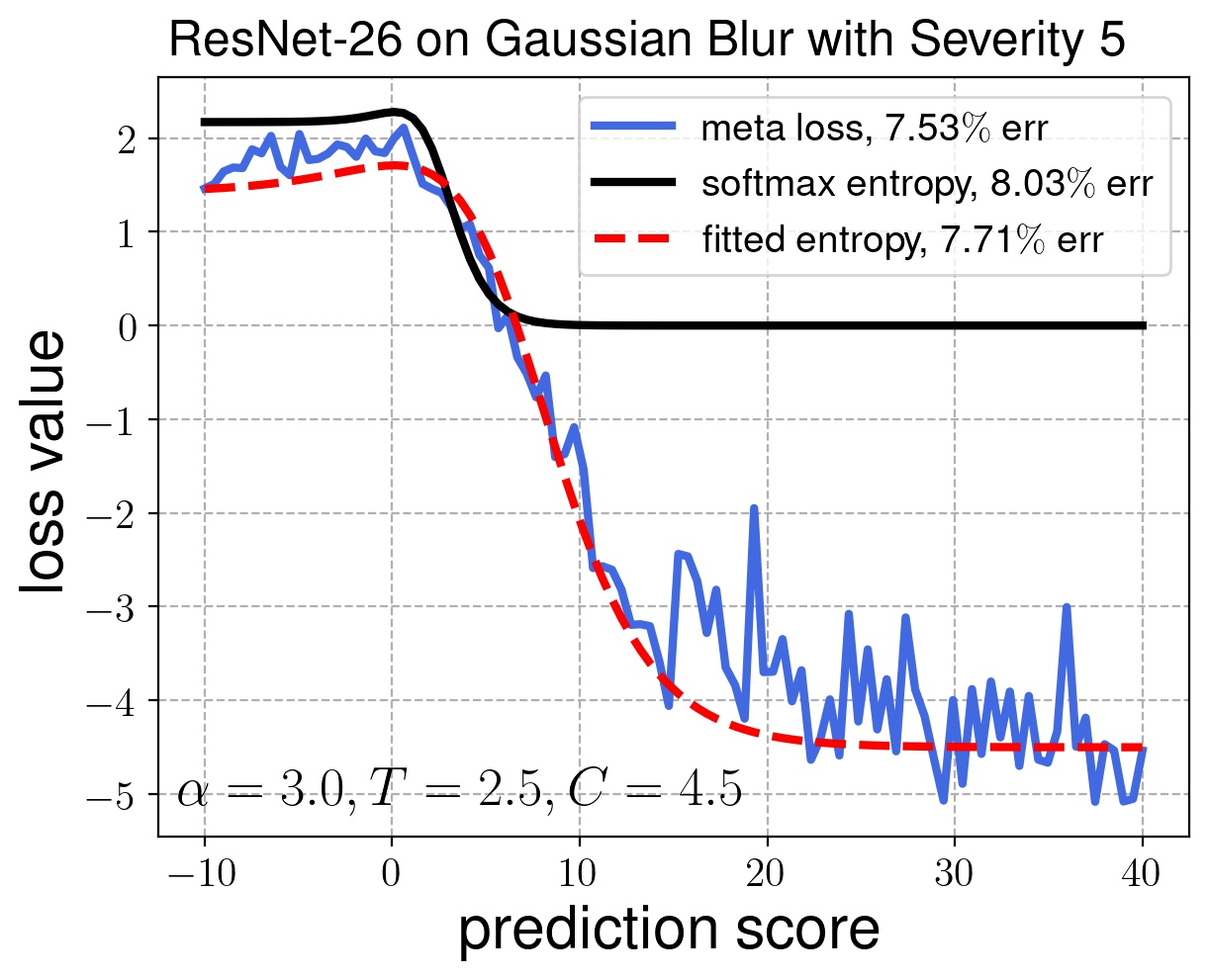}
% 		\vspace{-1ex}
		\caption{}
	\end{subfigure}
	\begin{subfigure}{0.40\textwidth}
		\includegraphics[width=\textwidth]{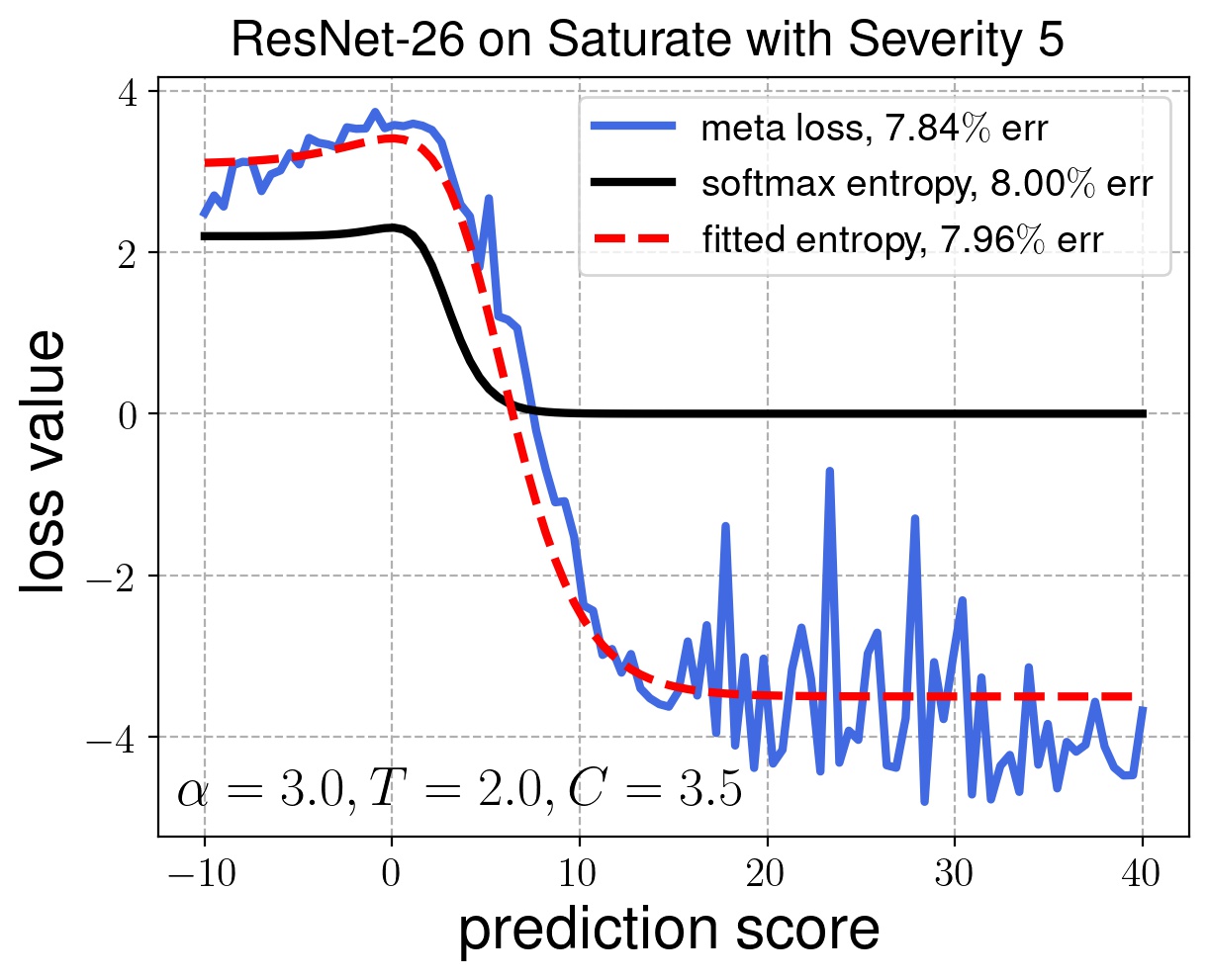}
% 		\vspace{-1ex}
		\caption{}
	\end{subfigure}
%%%%%%%%%%%%%%
	\begin{subfigure}{0.40\textwidth}
		\includegraphics[width=\textwidth]{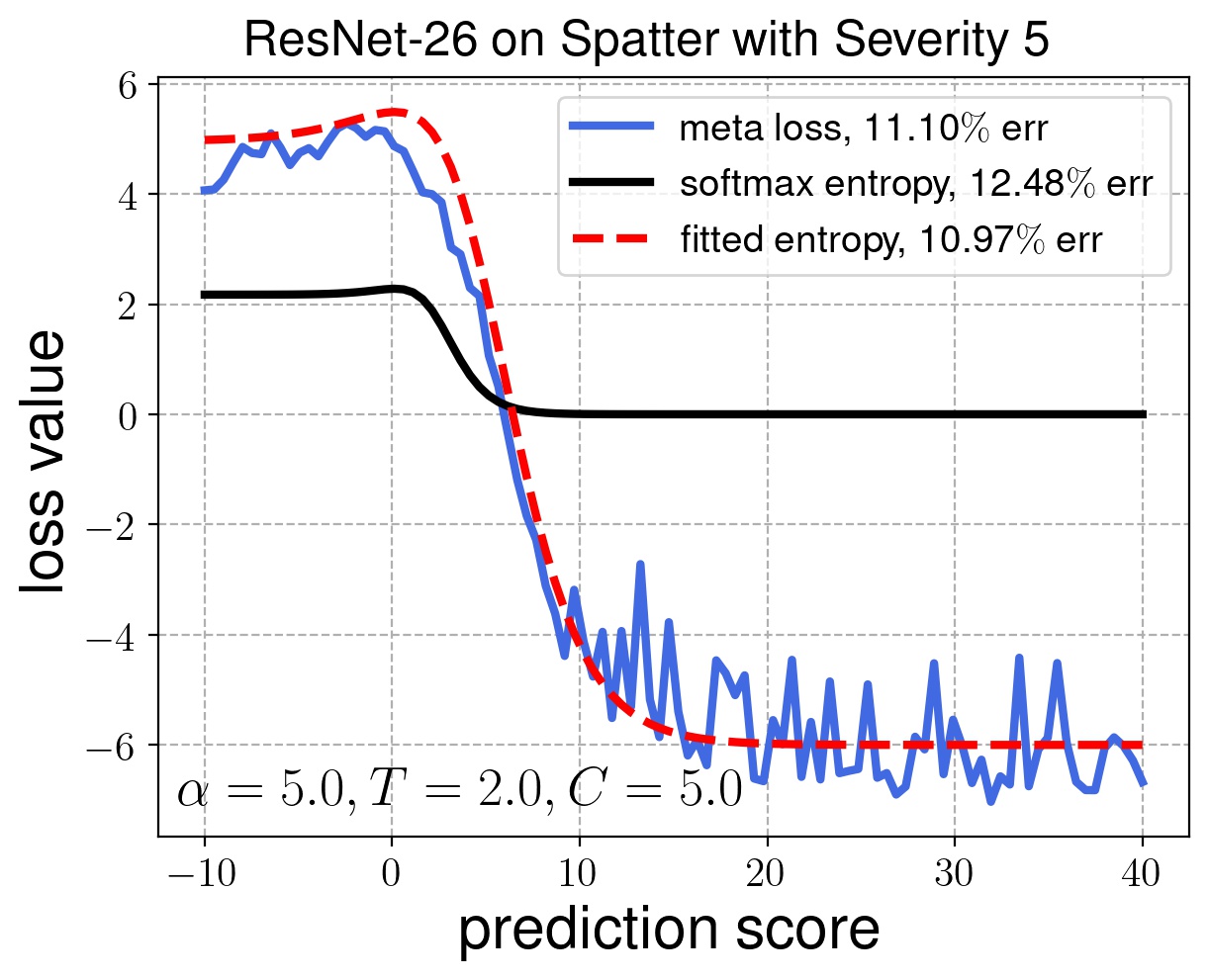}
% 		\vspace{-1ex}
		\caption{}
	\end{subfigure}
	\vspace{-1ex}
  \caption{Visualization of meta loss (blue) learnt from various noise types in CIFAR-10-C validation set, where base classifiers are trained with cross-entropy loss. We show the error of meta loss, softmax entropy and fitted entropy for test-time adaptation on the corresponding noise types. We also show the parameters ($\alpha, T, C$) in the fitted entropy.}
\label{appendix:resnet26_meta_loss_noise}
\end{figure}

\begin{figure}[!htbp]
	\centering
	\begin{subfigure}{0.40\textwidth}
		\includegraphics[width=\textwidth]{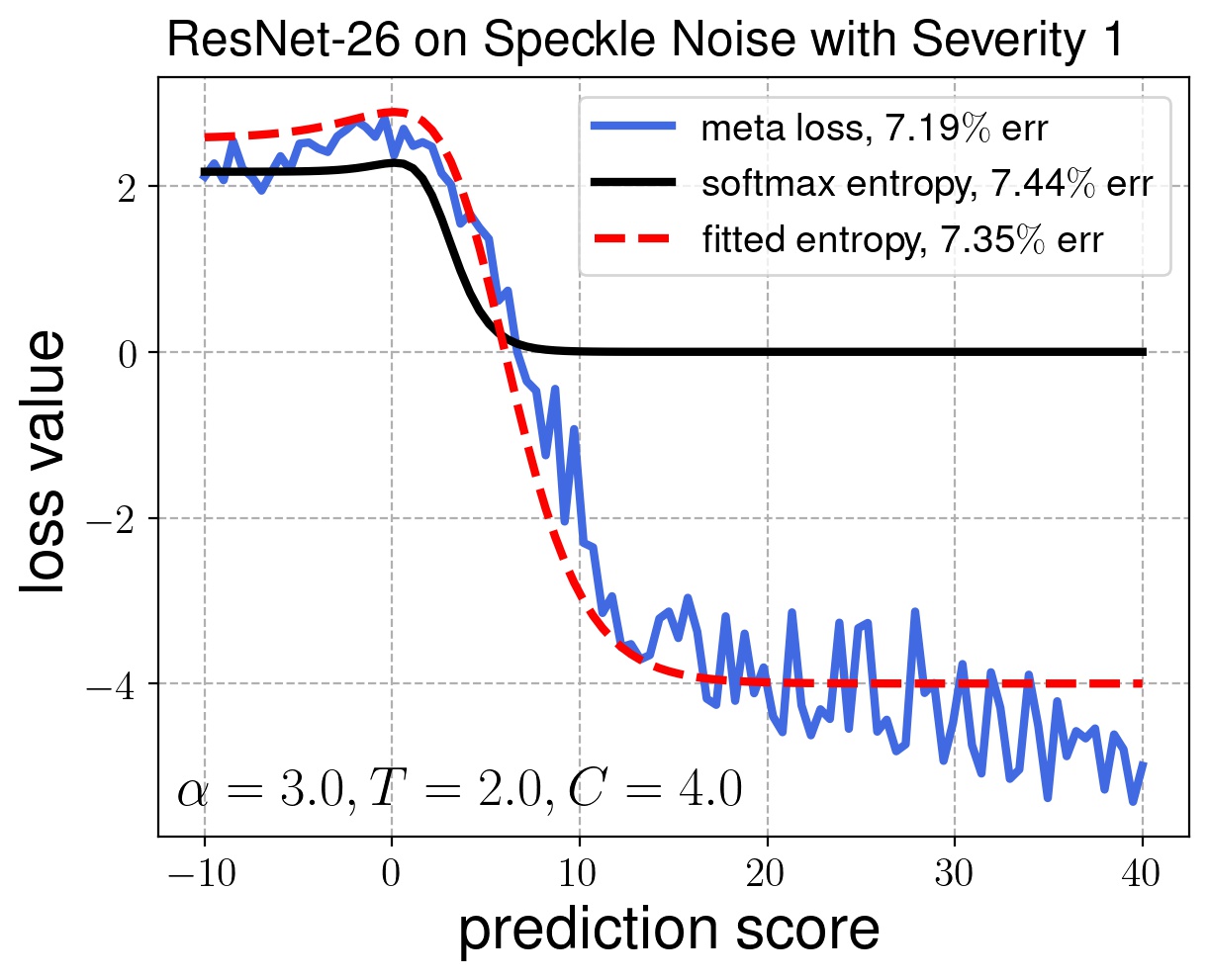}
% 		\vspace{-1ex}
		\caption{}
	\end{subfigure}
%%%%%%%%%%%%%%
	\begin{subfigure}{0.40\textwidth}
		\includegraphics[width=\textwidth]{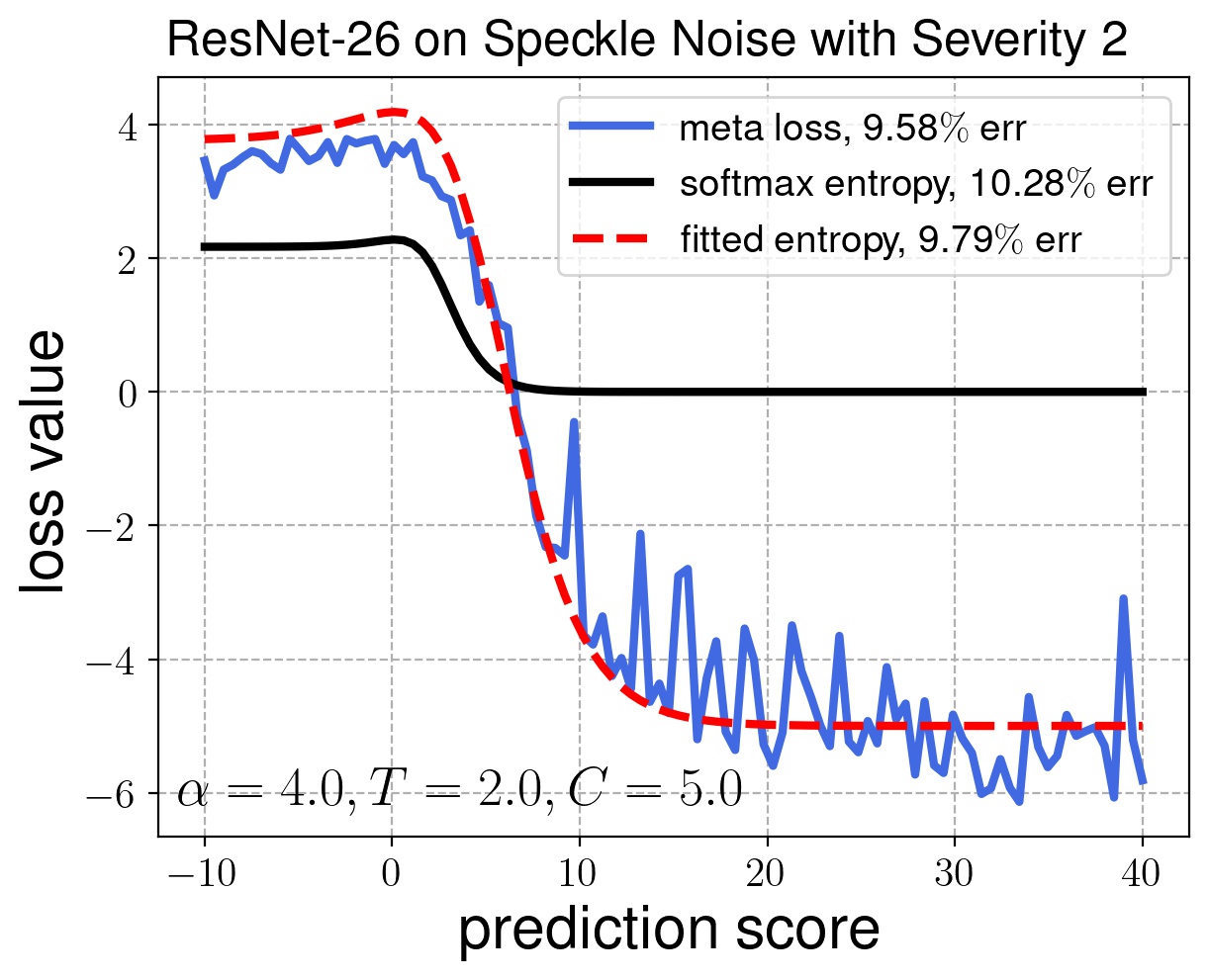}
% 		\vspace{-1ex}
		\caption{}
	\end{subfigure}
	\begin{subfigure}{0.40\textwidth}
		\includegraphics[width=\textwidth]{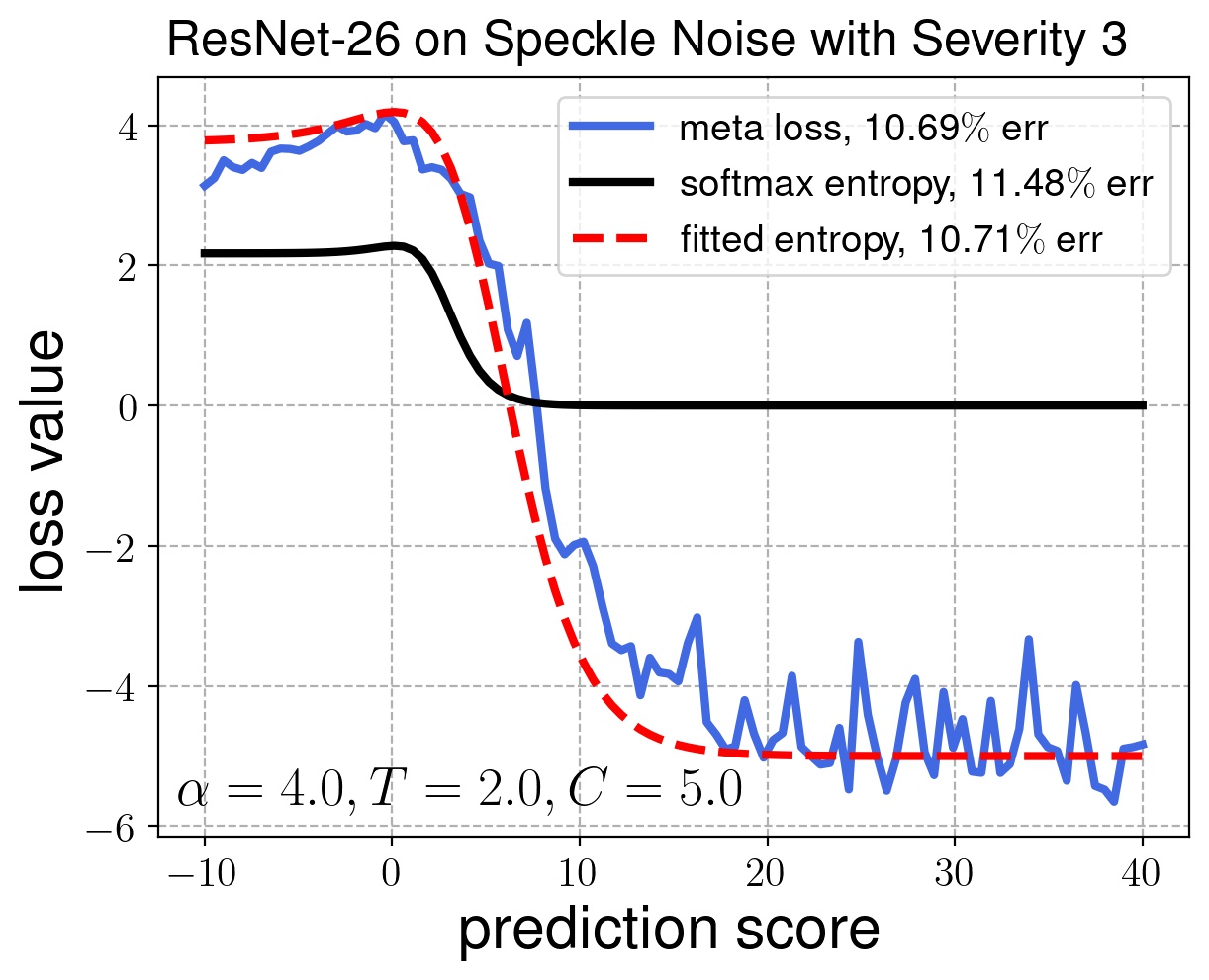}
% 		\vspace{-1ex}
		\caption{}
	\end{subfigure}
%%%%%%%%%%%%%%
	\begin{subfigure}{0.40\textwidth}
		\includegraphics[width=\textwidth]{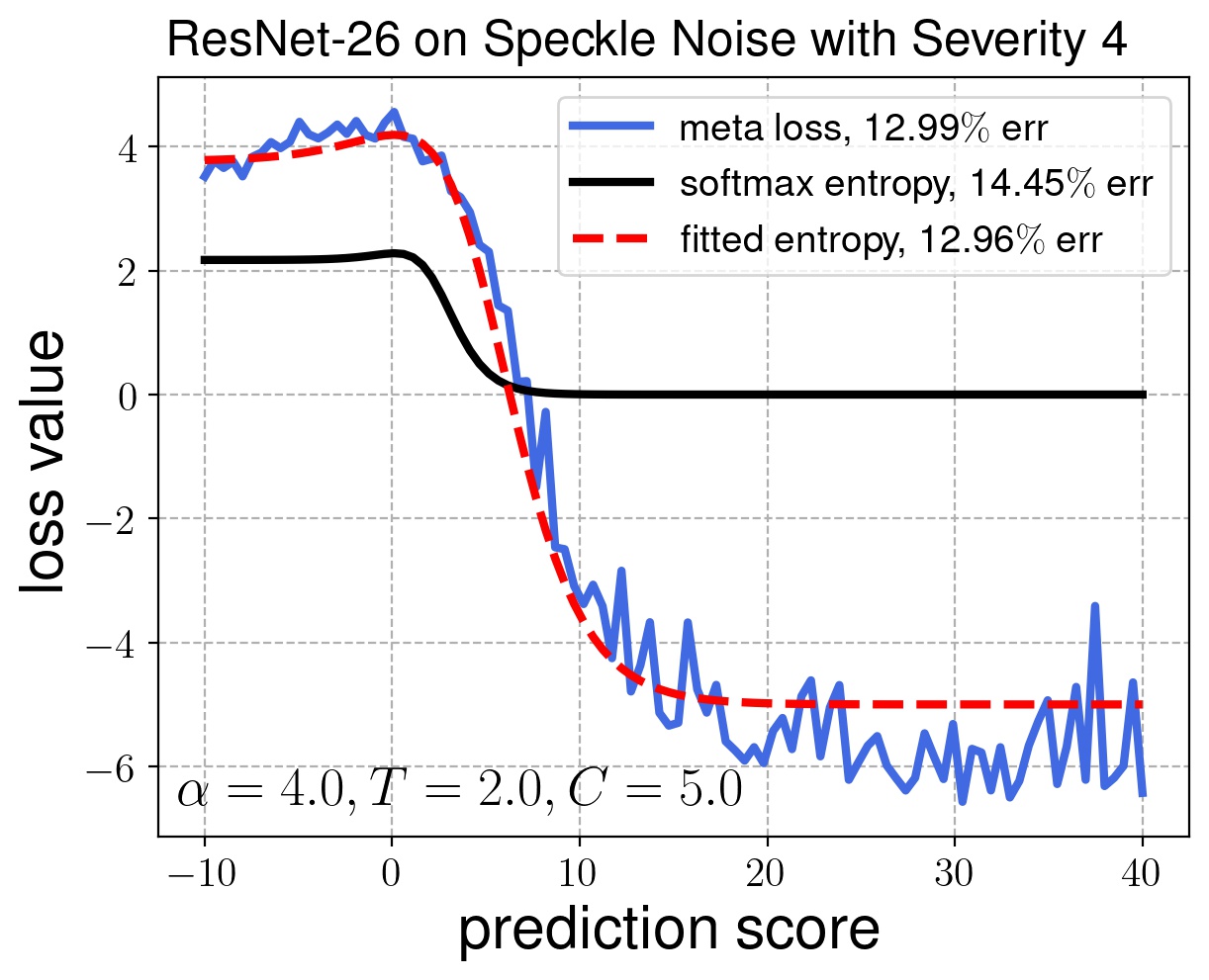}
% 		\vspace{-1ex}
		\caption{}
	\end{subfigure}
	\vspace{-1ex}
  \caption{Visualization of meta loss (blue) learnt on speckle noise with different severity level for CIFAR-10-C, where base classifiers are trained with cross-entropy loss. We show the error of meta loss, softmax entropy and fitted entropy for test-time adaptation on the corresponding noise types. We also show the parameters ($\alpha, T, C$) in the fitted entropy.}
\label{appendix:resnet26_meta_loss_sev}
\end{figure}

\begin{figure}[!htbp]
	\centering
	\begin{subfigure}{0.40\textwidth}
		\includegraphics[width=\textwidth]{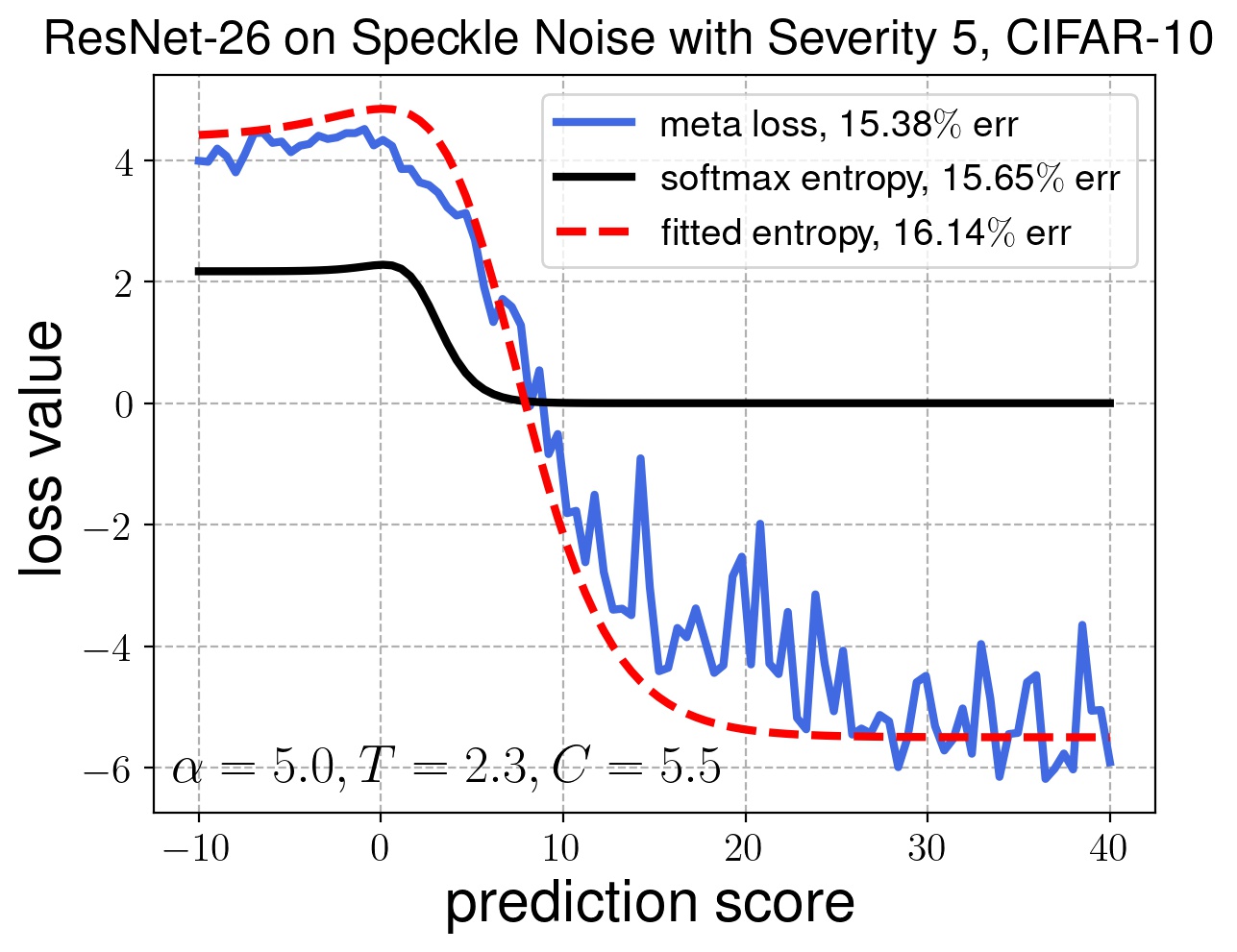}
% 		\vspace{-1ex}
		\caption{}
	\end{subfigure}
%%%%%%%%%%%%%%
	\begin{subfigure}{0.40\textwidth}
		\includegraphics[width=\textwidth]{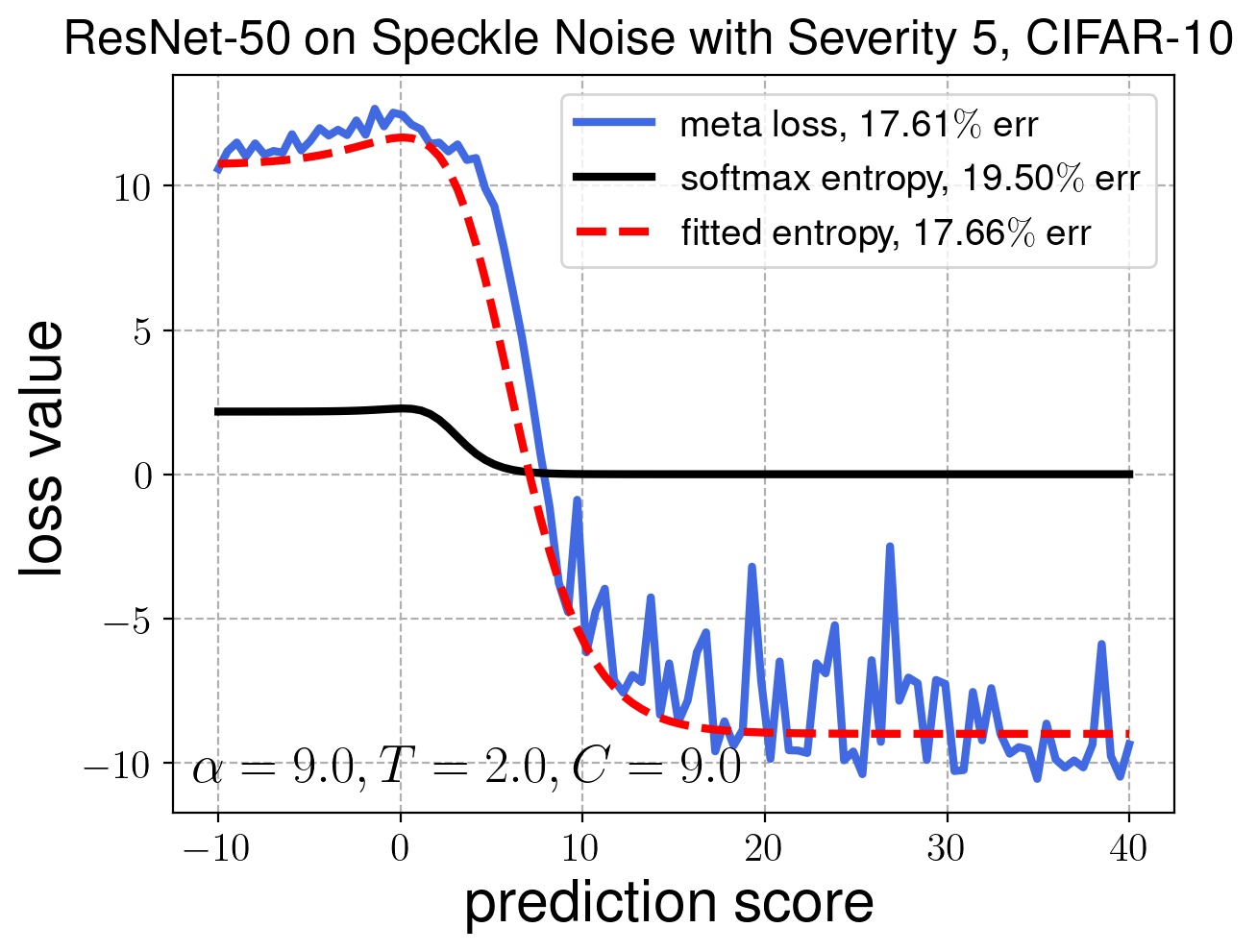}
% 		\vspace{-1ex}
		\caption{}
	\end{subfigure}
	\begin{subfigure}{0.40\textwidth}
		\includegraphics[width=\textwidth]{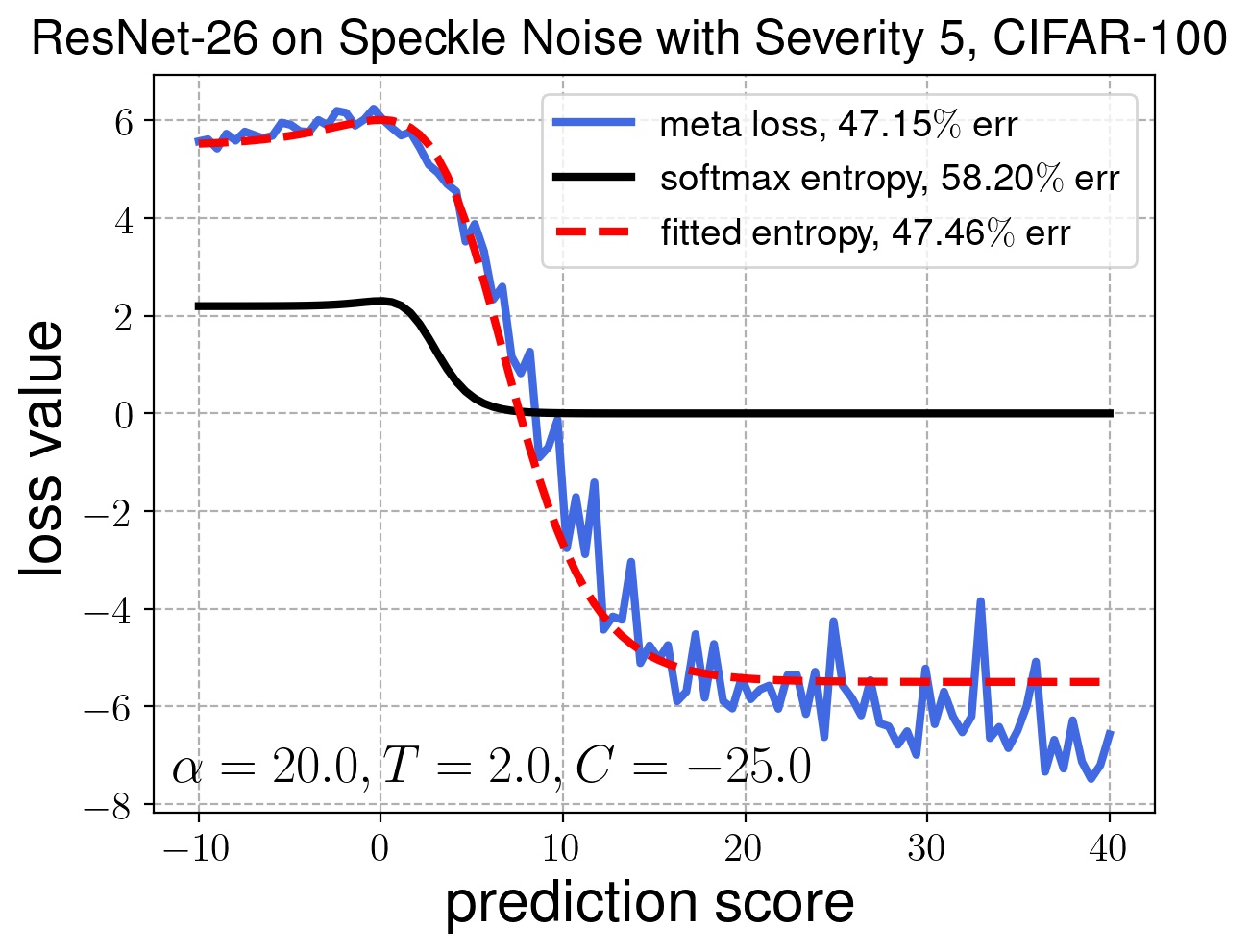}
% 		\vspace{-1ex}
		\caption{}
	\end{subfigure}
%%%%%%%%%%%%%%
	\begin{subfigure}{0.40\textwidth}
		\includegraphics[width=\textwidth]{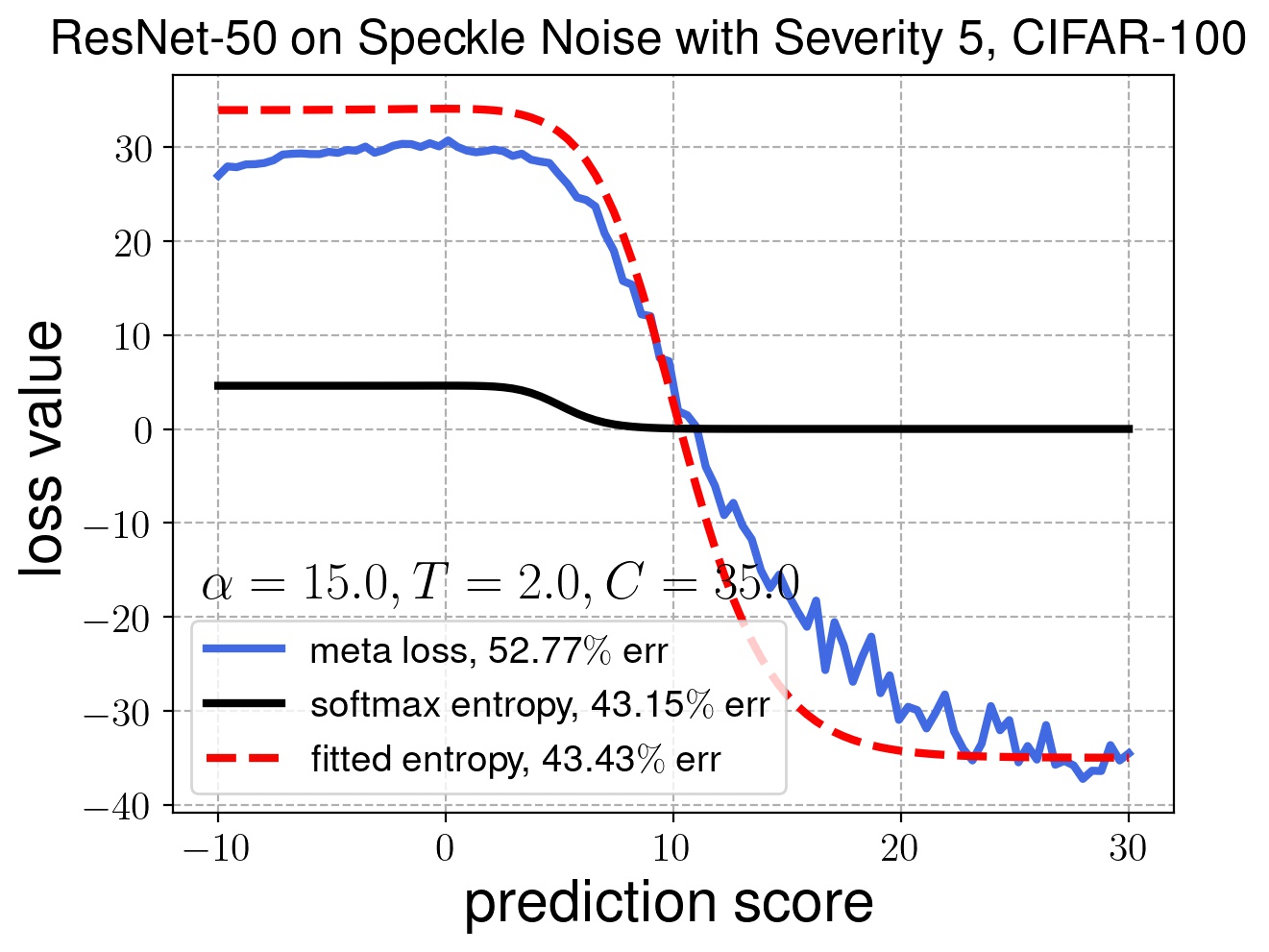}
% 		\vspace{-1ex}
		\caption{}
	\end{subfigure}
% 	\begin{subfigure}{0.40\textwidth}
% 		\includegraphics[width=\textwidth]{figures/meta_loss/imagenet_resnet50_speckle_sev5.jpg}
% % 		\vspace{-1ex}
% 		\caption{}
% 	\end{subfigure}
	\vspace{-1ex}
  \caption{Visualization of meta loss (blue) learnt across datasets (CIFAR-10-C/CIFAR-100-C) and base classifier architectures (ResNet-26/ResNet-50), where base classifiers are trained with cross-entropy loss. We show the error of meta loss, softmax entropy and fitted entropy for test-time adaptation on the corresponding noise types. We also show the parameters ($\alpha, T, C$) in the fitted entropy.}
\label{appendix:resnet26_meta_loss_dataset_arch}
\end{figure}

\begin{figure}[!htbp]
	\centering
%%%%%%%%%%%%%%
	\begin{subfigure}{0.40\textwidth}
		\includegraphics[width=\textwidth]{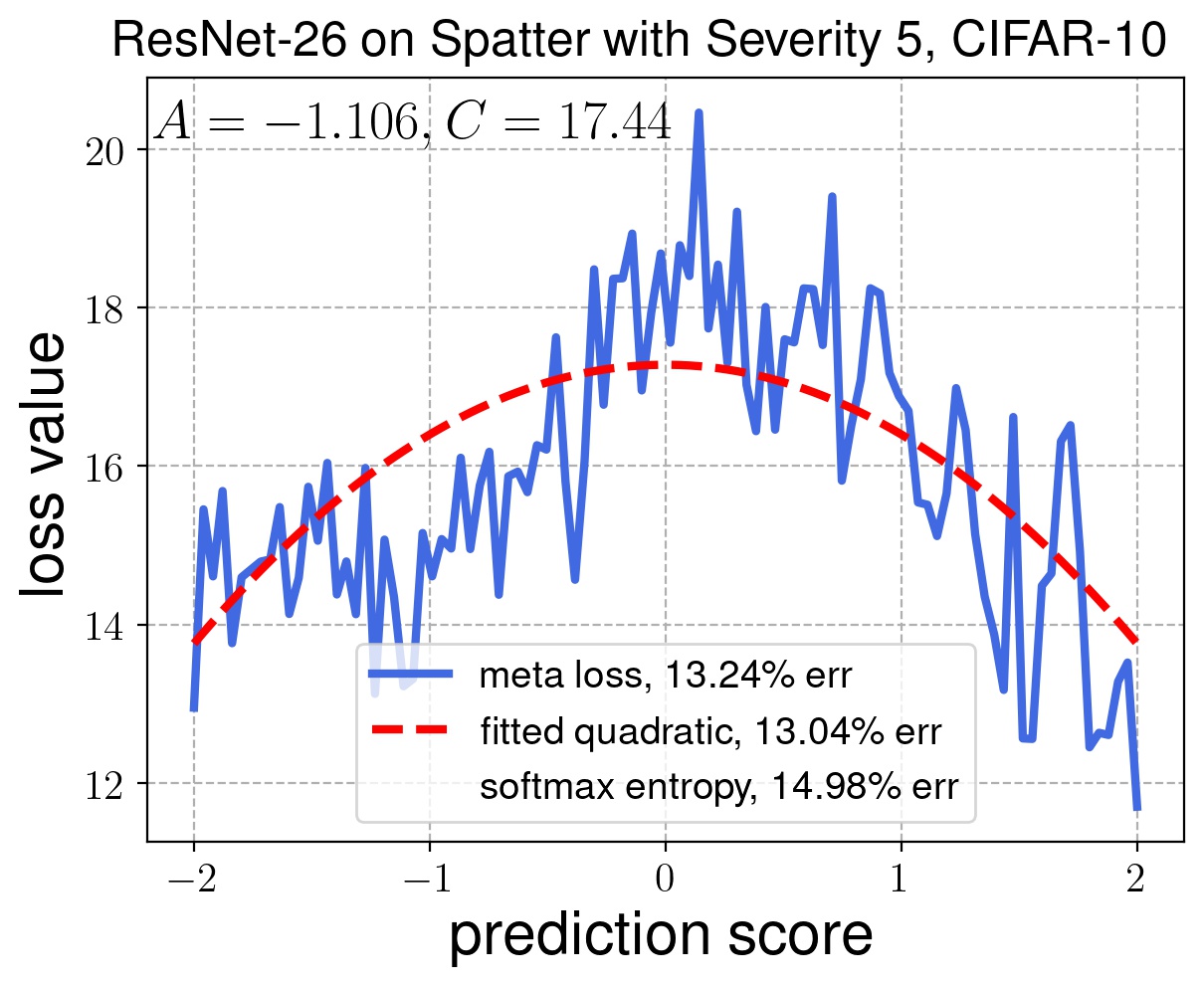}
% 		\vspace{-1ex}
		\caption{}
	\end{subfigure}
	\begin{subfigure}{0.40\textwidth}
		\includegraphics[width=\textwidth]{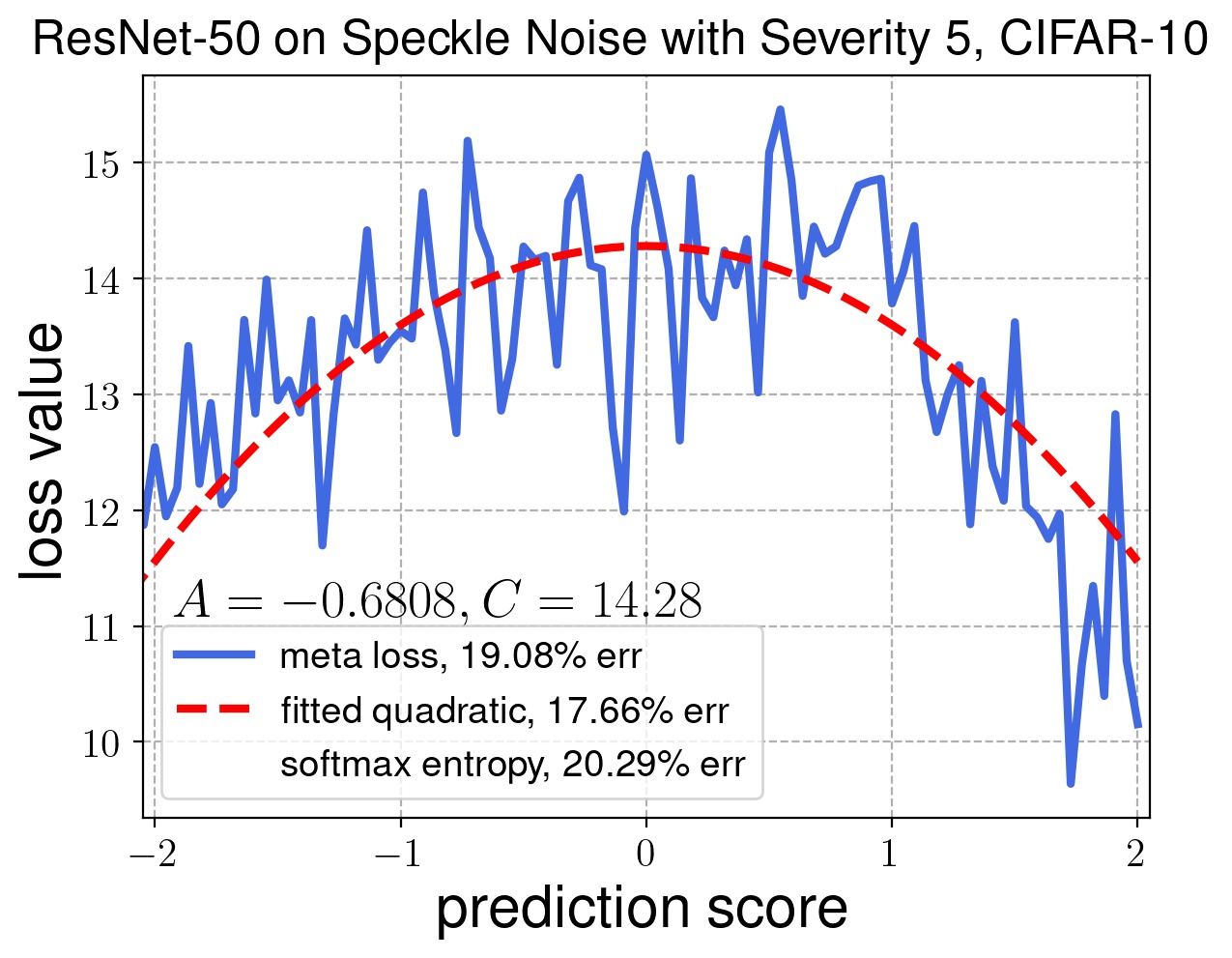}
% 		\vspace{-1ex}
		\caption{}
	\end{subfigure}
	\vspace{-1ex}
  \caption{Visualization of meta loss (blue), where base classifier is trained with quadratic loss. We show the error of meta loss, softmax entropy and fitted quadratic function for test-time adaptation on the corresponding noise types. We also show the parameters ($A, B, C$) in the fitted quadratic function.}
\label{appendix:resnet26_meta_loss_quad}
\end{figure}

%% file: tables/table-ce-hparams.tex
\begin{table}[!h]
    \centering
    \resizebox{0.9\textwidth}{!}{
\begin{tabular}{ccccc||cc}
    \toprule
    Dataset & \multicolumn{1}{c}{\begin{tabular}[c]{@{}c@{}}Temperature \\ (T)\end{tabular}}  & \multicolumn{1}{c}{Hard PL} & \multicolumn{1}{c}{Robust PL} & \multicolumn{1}{c||}{MEMO}  & \multicolumn{1}{c}{\begin{tabular}[c]{@{}c@{}}Conjugate PL\\ (\textbf{ENT})\end{tabular}}\\
    \midrule
    \multirow{2}{*}{CIFAR-10-C} &  \xmark & SGD,1$e^{-3}$, 1 & SGD,1$e^{-3}$, 1 & SGD,1$e^{-3}$, 1 & SGD, 1$e^{-3}$, 1 \\
    & \cmark  & SGD,1$e^{-3}$, 1 & SGD,1$e^{-2}$, 2 & SGD,5$e^{-3}$, 3 & Adam,1$e^{-3}$, 2 \\
    \midrule
    \multirow{2}{*}{CIFAR-100-C} &  \xmark & SGD,1$e^{-2}$, 1 & SGD,1$e^{-2}$, 1 & SGD,5$e^{-3}$, 1 & SGD, 1$e^{-2}$, 1 \\
    & \cmark  & SGD,1$e^{-2}$, 1 & SGD,1$e^{-2}$, 2 & SGD,1$e^{-2}$, 2 & SGD,1$e^{-2}$, 2 \\
    \midrule
    \multirow{2}{*}{ImageNet-C} & \xmark & SGD,1$e^{-2}$, 1 & SGD,2.5$e^{-3}$, 1 & SGD,1$e^{-3}$, 1 & SGD,2.5$e^{-3}$, 1 \\
    & \cmark  & SGD,1$e^{-2}$, 1 & SGD,2.5$e^{-3}$, 1.5 & SGD,1$e^{-3}$, 1 & SGD,2.5$e^{-3}$, 1.5 \\
    \bottomrule
    \end{tabular}
    }
    \vspace{1ex}
    \caption{Hyper-parameters (\textbf{Optimizer, Learning Rate, Temperature}) for the results in \autoref{tab:ce_loss}, where we showed the mean errors on the common corruptions dataset for a source classifier trained using cross-entropy loss.}
    \label{tab:ce_loss_hparams}
\end{table}
\vspace{-1em}

%% file: tables/table-polyloss1-hparams.tex
\begin{table}[!h]
    \centering
    \resizebox{1.0\textwidth}{!}{
\begin{tabular}{ccccccc||c}
    \toprule
    Dataset & T  & \multicolumn{1}{c}{Hard PL} & \multicolumn{1}{c}{Robust PL} & \multicolumn{1}{c}{ENT}  & \multicolumn{1}{c}{MEMO} & \multicolumn{1}{c||}{Softmax PL} & \multicolumn{1}{c}{\begin{tabular}[c]{@{}c@{}}Conjugate PL\\ (\textbf{Ours})\end{tabular}}\\
    \midrule
    \multirow{2}{*}{CIFAR-10-C} &  \xmark & SGD,1$e^{-3}$, 1 & SGD,1$e^{-3}$, 1 & SGD,1$e^{-3}$, 1 & SGD,5$e^{-3}$, 1 & SGD, 1$e^{-3}$, 1 & SGD, 1$e^{-3}$, 1 \\
    & \cmark  & SGD,1$e^{-3}$, 1 & SGD,1$e^{-2}$, 3 & SGD,1$e^{-2}$, 3 & SGD,5$e^{-3}$, 3 & SGD, 1$e^{-3}$, 2 & SGD, 1$e^{-3}$, 1.5 \\
    \midrule 
    \multirow{2}{*}{CIFAR-100-C} &  \xmark & SGD,1$e^{-2}$, 1 & SGD,1$e^{-2}$, 1 & SGD,1$e^{-2}$, 1 & SGD,1$e^{-2}$, 1 & SGD, 1$e^{-2}$, 1 & SGD, 1$e^{-2}$, 1 \\
    & \cmark  & SGD,1$e^{-2}$, 1 & Adam,1$e^{-3}$, 3 & SGD,1$e^{-2}$, 2 & SGD,1$e^{-2}$, 2 & SGD, 1$e^{-2}$, 2.5 & SGD, 1$e^{-2}$, 1.5 \\
    \midrule
    \multirow{2}{*}{ImageNet-C} &  \xmark & SGD,1$e^{-2}$, 1 & SGD,2.5$e^{-3}$, 1 & SGD,2.5$e^{-3}$, 1 & SGD,5$e^{-3}$, 1 & SGD, 2.5$e^{-3}$, 1 & SGD, 2.5$e^{-3}$, 1 \\
    & \cmark  & SGD,1$e^{-2}$, 1 & SGD,2.5$e^{-3}$, 1 & SGD,2.5$e^{-3}$, 1.5 & SGD,5$e^{-3}$, 1 & SGD, 2.5$e^{-3}$, 2 & SGD, 2.5$e^{-3}$, 1 \\
    \bottomrule
    \end{tabular}
    }
    \vspace{1ex}
    \caption{Hyper-parameters (\textbf{Optimizer, Learning Rate, Temperature}) for the results in Table~\ref{tab:polyloss1}, where we showed the mean errors on the common corruptions dataset for a source classifier trained using poly-loss.}
    \label{tab:poly_loss_hparams}
\end{table}
\vspace{-2em}

%% file: tables/table-quad-hparams.tex
\begin{table}[!t]
    \centering
    \resizebox{1.0\textwidth}{!}{
\begin{tabular}{ccccccc||c}
    \toprule
    Dataset & T  & \multicolumn{1}{c}{Hard PL} & \multicolumn{1}{c}{Robust PL} & \multicolumn{1}{c}{ENT}  & \multicolumn{1}{c}{MEMO} & \multicolumn{1}{c||}{Softmax PL} & \multicolumn{1}{c}{\begin{tabular}[c]{@{}c@{}}Conjugate PL\\ (\textbf{Ours})\end{tabular}}\\
    \midrule
    \multirow{1}{*}{CIFAR-10-C} &  \xmark & SGD,1$e^{-2}$, 1 & SGD,1$e^{-2}$, 1 & SGD,1$e^{-2}$, 1 & SGD,1$e^{-2}$, 1 & SGD,1$e^{-4}$, 1 & SGD,1$e^{-2}$, 1\\
    \midrule 
    \multirow{2}{*}{CIFAR-100-C} &  \xmark & Adam,1$e^{-3}$, 1 & Adam,1$e^{-3}$, 1 & Adam,1$e^{-3}$, 1 & Adam,1$e^{-3}$, 1 & Adam, 1$e^{-4}$, 1 & Adam, 1$e^{-3}$, 1 \\
    & \cmark  & Adam,1$e^{-3}$, 1 & Adam,1$e^{-3}$, 0.5 & Adam,1$e^{-3}$, 2 & Adam,1$e^{-3}$, 2 & Adam, 1$e^{-4}$, 2.5 & Adam, 1$e^{-3}$, 1 \\
    \bottomrule
    \end{tabular}
    }
    \vspace{1ex}
    \caption{Hyper-parameters (\textbf{Optimizer, Learning Rate, Temperature}) for the results in Table~\ref{tab:squaredloss_results}, where we showed the mean errors on the common corruptions dataset for a source classifier trained using squared loss.}
    \label{tab:squared_loss_hparams}
\end{table}
\vspace{-2em}

%% file: tables/table-ce-digit.tex
\begin{table}[h]
    \centering
    \resizebox{0.9\textwidth}{!}{
\begin{tabular}{ccccc||cc}
    \toprule
    Dataset & \multicolumn{1}{c}{\begin{tabular}[c]{@{}c@{}}Temperature \\ (T)\end{tabular}}  & \multicolumn{1}{c}{Hard PL} & \multicolumn{1}{c}{Robust PL} & \multicolumn{1}{c||}{MEMO}  & \multicolumn{1}{c}{\begin{tabular}[c]{@{}c@{}}Conjugate PL\\ (\textbf{ENT})\end{tabular}}\\
    \midrule
    \multirow{2}{*}{SVHN $\rightarrow$ MNIST} &  \xmark & 21.54 & 27.44 & \textbf{10.67} & 14.41 & \\
    & \cmark  & 21.54 & 13.26 & \textbf{9.36} & \textbf{9.26} & \\
    \midrule
    \multirow{2}{*}{SVHN $\rightarrow$ USPS} &  \xmark & 26.06 & 26.81 & 22.72 & \textbf{22.57} & \\
    & \cmark   & 26.06 & \textbf{22.32} & 22.42 & \textbf{22.27} & \\
    \bottomrule
    \end{tabular}
    }
    \vspace{1ex}
    \caption{Mean errors when adapting to digit adaptation benchmarks using a source classifier trained via cross-entropy loss. Here, conjugate pseudo-labeling becomes softmax-entropy minimization. Again we observe that with the right temperature scaling, softmax-entropy minimization matches other approaches. For additional context, the source classifier errors without adaptation are: SVHN $\rightarrow$ MNIST ($34.17\%$), SVHN $\rightarrow$ USPS ($31.84\%$).}
    \label{tab:ce_loss_digit}
\end{table}
\vspace{-1em}